\documentclass{article}

\usepackage{arxiv}

\usepackage[utf8]{inputenc} 
\usepackage[T1]{fontenc}    
\usepackage{hyperref}       
\usepackage{url}            
\usepackage{booktabs}       
\usepackage{amsfonts}       
\usepackage{nicefrac}       
\usepackage{microtype}      
\usepackage{lipsum}		
\usepackage{graphicx}
\usepackage{natbib}
\usepackage{doi}
\usepackage{geometry}
\usepackage{multirow}
\usepackage{booktabs}
\usepackage{xcolor}
\usepackage{tabularray}
\usepackage{lscape} 
\usepackage{rotating}
\usepackage{tabularx}
\usepackage[labelfont=bf]{caption} 

\usepackage{ragged2e}
\usepackage{lipsum}

\title{Evaluating the Role of Data Enrichment Approaches Towards Rare Event Analysis in Manufacturing}

\author{ \href{https://orcid.org/0000-0002-5320-5566}{\includegraphics[scale=0.06]{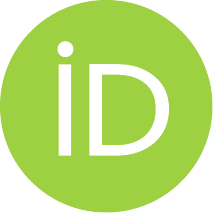}\hspace{1mm}Chathurangi Shyalika}\\
	AI Institute, University of South Carolina \\
	USA \\
	\texttt{jayakodc@email.sc.edu} \\
	\And
	\href{https://orcid.org/0000-0001-5810-1849}{\includegraphics[scale=0.06]{orcid.pdf}\hspace{1mm}Ruwan Wickramarachchi} \\
	AI Institute, University of South Carolina \\
	USA \\
	\texttt{ruwan@email.sc.edu} \\
 \And
	\href{https://orcid.org/0000-0003-1189-0291}{\includegraphics[scale=0.06]{orcid.pdf}\hspace{1mm}Fadi El Kalach} \\
	McNair Center \\Department of Mechanical Engineering \\University of South Carolina \\
	USA \\
	\texttt{elkalach@email.sc.edu} \\
  \And
	\href{https://orcid.org/0000-0003-1452-9653}{\includegraphics[scale=0.06]{orcid.pdf}\hspace{1mm}Ramy Harik} \\
	McNair Center \\Department of Mechanical Engineering \\University of South Carolina \\
	USA \\
	\texttt{harik@mailbox.sc.edu} \\
  \And
 \href{https://orcid.org/0000-0002-0021-5293}{\includegraphics[scale=0.06]{orcid.pdf}\hspace{1mm}Amit Sheth} 
 \\
 AI Institute \\
  University of South Carolina \\
  USA\\
  \texttt{amit@sc.edu} \\
}

\renewcommand{\shorttitle}{\textit{arXiv} Template}

\hypersetup{
pdftitle={A template for the arxiv style},
pdfsubject={q-bio.NC, q-bio.QM},
pdfauthor={David S.~Hippocampus, Elias D.~Striatum},
pdfkeywords={First keyword, Second keyword, More},
}

\begin{document}

\maketitle


\begin{abstract}
Rare events are occurrences that take place with a significantly lower frequency than more common regular events. In manufacturing, predicting such events is particularly important, as they lead to unplanned downtime, shortening equipment lifespan, and high energy consumption. The occurrence of events is considered frequently-rare if observed in more than 10\% of all instances, very-rare if it is 1-5\%, moderately-rare if it is 5-10\%, and extremely-rare if less than 1\%. The rarity of events is inversely correlated with the maturity of a manufacturing industry. Typically, the rarity of events affects the multivariate data generated within a manufacturing process to be highly imbalanced, which leads to bias in predictive models. This paper evaluates the role of data enrichment techniques combined with supervised machine-learning techniques for rare event detection and prediction. To address the data scarcity, we use time series data augmentation and sampling methods to amplify the dataset with more multivariate features and data points while preserving the underlying time series patterns in the combined alterations. Imputation techniques are used in handling null values in datasets. Considering 15 learning models ranging from statistical learning to machine learning to deep learning methods, the best-performing model for the selected datasets is obtained and the efficacy of data enrichment is evaluated. Based on this evaluation, our results find that the enrichment procedure enhances up to 48\% of F1 measure in rare failure event detection and prediction of supervised prediction models.  We also conduct empirical and ablation experiments on the datasets to derive dataset-specific novel insights. Finally, we investigate the interpretability aspect of models for rare event prediction, considering multiple methods.
\footnote{Code and data to reproduce the results are available at this \href{https://anonymous.4open.science/status/Rare_Event_Analysis-B5E6}{link}}.

\end{abstract}


\section{Introduction}
Events are occurrences linked to particular locations (spatial), time frames (temporal), and contexts (semantics). Rare events, a subset of these, are notable for their infrequency. The degree of the infrequency of rare events is typically influenced by the specific field of application (\cite{harrison2016rare, glasserman1999multilevel}). Rare events can be categorized into distinct categories based on factors like the distribution of data, significant differences in rarity levels, and context: extremely rare with a frequency of 0-1\%, very rare with a frequency of 1-5\%, moderately rare with a frequency of 5-10\%, and frequently-rare with a frequency greater than 10\% \cite{shyalika2023comprehensive}. In real life, rare events can be observed ubiquitously in various industries such as earth science, manufacturing, telecommunication, healthcare, transportation, economy, and energy.
They span various applications, including rare disease diagnosis, anomaly detection, fraud detection, and natural disaster prediction (\cite{shyalika2023comprehensive}). Rare events, notable for their scarcity, provide vital information in any domain and pose challenges including but not limited to the "Curse of Rarity" (CoR) (\cite{liu2022curse}). This leads to hindering the decision-making, modeling, verification, and validation due to their exceptional rarity (\cite{harrison2016rare, glasserman1999multilevel, omar2022exploring, liu2022curse, meng2022empirical, jalayer2021fault, yan2019deep, souza2021feature}). Rare events, by their intrinsic nature, exhibit significant challenges in the manufacturing domain. Some examples of rare events in manufacturing are paper breaks in the paper manufacturing industry, component failures in the automotive industry, supply chain disruptions in food processing and packaging, etc. Rare events can have a significant impact when they occur. These disruptions are costly for industries; resumption time is higher. For example, in pulp-and-paper manufacturing, paper breakage that occurs <1\%, can cost the industry a huge loss (\cite{ranjan2018dataset}). Predicting such hard-to-predict occurrences is important for cost management, operational efficiency, and energy conservation.  Identifying them leads to reducing defects, lowering equipment downtime, and optimizing energy consumption. Thus, the identification and anticipation of such events hold paramount importance for ensuring the overall productivity of the industry. The sporadic occurrence and often unpredictable nature of these events make their detection and prediction imperative. While detecting such events is significant, the ability to foresee them beforehand assumes greater importance, facilitating proactive measures to mitigate potential consequences. It would lead to ensuring optimization, quality, and safety standards in manufacturing processes. 

There is a difference between the detection and prediction of rare events in any industry based on their occurrence. Detection involves identifying the occurrence of rare events after they arise, whereas prediction involves forecasting the likelihood of future occurrences of such events before they manifest. Detection emphasizes recognizing and flagging rare occurrences within use cases that deviate significantly from the norm or occur infrequently. Prediction revolves around anticipating, forecasting, or modeling the occurrence of infrequent events based on historical data and patterns.

In machine learning, the quality and richness of data play an integral role in the development and efficacy of predictive models. Data enrichment techniques emerge as pivotal methodologies employed to augment and refine raw datasets, infusing them with additional information, context, and structure. The core concept of data enrichment is to address the deficiencies present in raw datasets, including missing data points (incompleteness), scarce representations (sparsity), and underlying distortions or prejudices (inherent biases). Data enrichment benefits machine learning models, imparting them with enhanced capabilities to comprehend complex patterns, generalize effectively, and make more accurate predictions.

Common domain-agnostic data enrichment techniques include data augmentation, sampling, imputation, domain knowledge integration, data transformation, etc. Our study particularly focuses on data augmentation, sampling, and imputation. Data augmentation focuses on generating synthetic data to increase the input space (\cite{wen2020time, semenoglou2023data, iglesias2022data}). In our study, we focused on increasing only the feature space of the data while keeping the number of data points across the time steps unchanged. Sampling techniques aim to balance the data distribution by synthesizing or eliminating the number of data points, which is a common technique used in imbalanced data analysis (\cite{mohammed2020machine}). Imputation techniques focus on creating accurate and complete data by making adjustments for unspecified values in the data (\cite{lakshminarayan1996imputation, jerez2010missing}). The amalgamation of these techniques offers an opportunity to leverage multivariate information and compensate for the scarcity of event instances. These would overall lead to enhancing the quality, diversity, and utility of the dataset for machine learning tasks.

To the best of our knowledge, none of the previous studies explored the potential of applying different data enrichment techniques using different classifiers in rare-event prediction on real-world manufacturing datasets. This research investigates the application of statistical, machine learning, and deep learning-based modeling techniques with three data resampling techniques, 11 data augmentation techniques, and two data imputation techniques in predicting rare events. 

The main contributions of this paper are listed below:

\subsection{Primary contributions:}
\begin{enumerate}
    \item The primary contribution of this work is to propose a framework to investigate the role of data enrichment approaches in rare-event detection and prediction. (Data augmentation, Sampling and Imputation)

    \item Introducing a real-world dataset from a product assembly manufacturing industry.
\end{enumerate}

\subsection{Secondary contributions:}
\begin{enumerate}
    \item Conduct empirical and ablation experiments on five real-world datasets from the manufacturing domain for rare event detection and prediction to derive dataset-specific novel insights.

    \item Investigate the interpretability aspect of models for rare event prediction considering multiple methods.
\end{enumerate}

The rest of the paper is organized as follows: the related work section presents previous studies on leveraging data enrichment techniques in rare event detection and prediction. The methodology section provides a brief overview of the datasets adopted, methods used for rare event detection and prediction, and data enrichment approaches used in this study. The modeling results with experiments have been included in the evaluation section. It discusses the results of the proposed models and model and feature analysis based on explainable learning methods. Finally, the conclusion and future research section provide an overall summary of this research along with significant findings, limitations, and future scope of research.

\section{Related works}

In early studies, data enrichment techniques have been used to enhance the existing datasets with additional information to improve the quality, diversity, and predictive power. Considering the scope of our research, we explore the related work across three data augmentation techniques: data augmentation, sampling, and imputation. Data augmentation techniques expand dataset size and diversity, sampling methods address class distribution balance, and imputation strategies effectively manage null values within datasets. Data augmentation is a widely employed method in machine learning that involves expanding the size and diversity of datasets by creating additional data samples through various transformations while retaining the original labels. 
Time series data augmentation has received considerable attention in the literature (\cite{bandara2021improving, wen2020time}). With respect to rare event prediction, these techniques exhibit utility across a wide array of domains, including machinery fault diagnosis, computer vision, and geology. For instance, in a manufacturing-related use case (\cite{Ranjan2019DataCD}), researchers introduced data augmentation by generating new features through Fast Fourier Transform (FFT) alongside given features, substantially enhancing a multivariate time series dataset (\cite{ranjan2018dataset}) in pulp-and-paper manufacturing. Other studies in rare event prediction have leveraged data augmentation techniques, including variations of Generative Adversarial Networks (GAN), conditional GAN (CGAN), and Wasserstein GAN (WGAN) (\cite{fathy2020learning}). These approaches have generated labeled samples for various applications, such as mineral prospectivity prediction and scene change event detection, using image-based data augmentation techniques like cropping geological image data and window-based augmentation (\cite{yang2022applications, hamaguchi2019rare, parsa2021deep}). These methods contribute to enhancing the robustness and effectiveness of rare event prediction models by enriching the available datasets.

Sampling techniques are integral to machine learning, especially when handling imbalanced datasets, such as those focused on anomalies, failures, and rare events. Rare event research has employed various categorizations for sampling methods, including both basic and advanced methods. Basic sampling techniques involve randomly undersampling the majority class or oversampling the minority class. In rare event research, these methods have been applied to balance the class distributions (\cite{seiffert2007mining, zhao2018framework, ranjan2018dataset, jo2004class, wu2007local, ahmadzadeh2019rare}). 
Researchers have explored combining sampling techniques with clustering models (\cite{nugraha2020clustering, wu2007local}), ensemble learning methods (\cite{chen2004using}), advanced architectures like Siamese CNN (\cite{hamaguchi2019rare}), and statistical sampling methods like Hoeffding bounds (\cite{iyer2015statistical}) to enhance predictive performance in rare event prediction. Yet, random sampling has drawbacks, including overfitting in oversampling and data loss in undersampling. Advanced sampling techniques, on the other hand, delve into more sophisticated strategies to intelligently balance classes. Widely applied methods include Synthetic Minority Oversampling Technique (SMOTE) (\cite{zhao2018framework, li2017adaptive, ali2014dynamic, ashraf2023identification, fathy2020learning, seiffert2007mining}), which generates synthetic minority cases to address overfitting, and Adaptive Synthetic Sampling (ADASYN) (\cite{he2008adasyn, ashraf2023identification}), which provides weighted oversampling to tackle difficulty in learning. 
For example, it has been observed that logistic regression, when coupled with SMOTE, has yielded better outcomes in the identification of Look-Alike-Sound-Alike (LASA) cases in textual data (\cite{zhao2018framework}). Furthermore, Asraf et al. (\cite{ashraf2023identification}) have leveraged ADASYN with the XGBoost model to facilitate rare event modeling in identifying high-risk segments in Wrong-Way Driving.

Other advanced sampling techniques used in rare event research include Edited Nearest Neighbor (ENN) (\cite{tomek1976experiment, li2017rare, ashraf2023identification}), Neighborhood Cleaning Rule (NCL) (\cite{bekkar2013imbalanced, li2017rare}), NearMiss (NM)(\cite{mani2003knn, li2017rare, ashraf2023identification}), and One-Sided Selection (OSS)(\cite{hart1968condensed, kubat1997addressing, kubat1998machine, chen2004using, seiffert2007mining}), which focus on reducing noise and refining decision boundaries. 
One-sided Selection (OSS) is an undersampling technique that intelligently removes noisy majority-class instances, using Tomek Links (TL) and the Condensed Nearest Neighbor (CNN) rule (\cite{hart1968condensed}), to create a more refined and representative subset of the majority class in rare event studies (\cite{kubat1997addressing, kubat1998machine, chen2004using, seiffert2007mining}). TL can remove noise and boundary points in majority class samples in rare events (\cite{kotsiantis2006handling, li2017rare, ashraf2023identification, kubat1998machine}). Furthermore, advanced methods such as Cluster-based Oversampling (CBO) handle both between-class and within-class imbalances simultaneously (\cite{jo2004class, seiffert2007mining}). Time Series subsampling reduces the computational cost of analyzing long-time series data by selecting a subset of time points from the original to create a new, shorter, and more balanced time series (\cite{fukuchi1999subsampling, combes2022time}). Stratified sampling ensures a proportional representation of classes in the sampled dataset by dividing the dataset into subgroups (or \textit{strata}) based on the target variable's classes and then randomly sampling from each stratum. 
In industrial machinery fault diagnosis (\cite{liu2021machinery}), subsampling of time series data was employed to ensure a balanced representation of normal and different bearing fault types; also, it uses stratified techniques to appropriately represent these subgroups in the final dataset. Additionally, audio data sampling (\cite{abbasi2022large}), uncertainty sampling (\cite{pickering2022discovering}), and choice-based or endogenous sampling  (\cite{maalouf2018logistic, van2006prediction}) cater to specific data modalities and sampling goals, making them valuable in rare event prediction use cases.

Imputation techniques play a pivotal role in addressing missing or incorrect data values in datasets, ensuring their completeness and accuracy. These techniques can be categorized into two main types: simple imputation and advanced imputation. Simple imputation methods encompass straightforward strategies such as mean or median substitution, which replace missing values with the mean or median of the respective attribute (\cite{xiu2021variational, rafsunjani2019empirical, fathy2020learning, gondek2016prediction}). Interpolation is another simple imputation approach that estimates missing values based on neighboring observed data points, considering underlying data trends (\cite{radi2015estimation, ahmadzadeh2019rare}). These simple techniques have been used in predicting cancer survival among adults in SEER dataset (\cite{xiu2021variational}), predicting failures in Air Pressure System (APS) dataset (\cite{rafsunjani2019empirical, fathy2020learning, gondek2016prediction}), and solar flare forecasting (\cite{ahmadzadeh2019rare}) for the imputation of missing values in rare-event time series prediction.

In contrast, advanced imputation methods delve into more complex approaches such as iterative imputation, multiple imputation, soft impute, expectation maximization, offset value approximation, and Singular Value Decomposition (SVD) imputation. These techniques provide a more sophisticated means of estimating missing values, utilizing statistical modeling and iterative processes. Advanced imputation approaches are particularly valuable in rare event prediction, as they consider intricate data patterns and enhance the overall quality of datasets. Researchers have applied a variety of imputation methods in the context of rare event prediction, yielding improved model performance and predictions (\cite{adil2022deep,omar2022exploring, rafsunjani2019empirical, yao2018accelerated, do2008expectation, wei2018missing, troyanskaya2001missing, cheon2009bayesian}).
For instance, Omar et al. (\cite{omar2022exploring}) employed the MisForest algorithm (\cite{stekhoven2012missforest}), which is an iterative imputation method based on the Random Forest algorithm for the imputation of missing data. Cheon et al. (\cite{cheon2009bayesian}) utilized advanced offset value approximation techniques involving predefined offset values for input data. Furthermore, Rafsunjani et al. (\cite{rafsunjani2019empirical}) conducted an empirical investigation of various imputation techniques, including expectation maximization, mean imputation, soft impute, Multiple Imputation by Chained Equation (MICE), and Iterative SVD, in the context of APS failure prediction, highlighting MICE as a high-performance method in their predictions.

From the studied literature, it was seen that even though much research has employed data enrichment techniques in multiple domains, there's still a lack of applicability in the manufacturing domain. 
Given the scarcity, complexity, multidimensionality, and heterogeneity of manufacturing data (\cite{ismail2019manufacturing}), it is imperative to explore methods for improving predictive performance. Data enrichment techniques can play a significant role in this context. Exploring the synergistic potential of combining various data enrichment techniques presents a promising avenue for addressing the above rare event prediction challenges. This innovative approach represents a novel direction that has yet to be extensively explored within the field. By strategically utilizing these robust methodologies, we can harness the complete capabilities of the data and attain heightened levels of precision in manufacturing-based predictive analyses.

\section{Methodology}
\label{sec:Methodology}
\begin{figure}[!ht]
  \centering
   \includegraphics[width=0.5\linewidth]{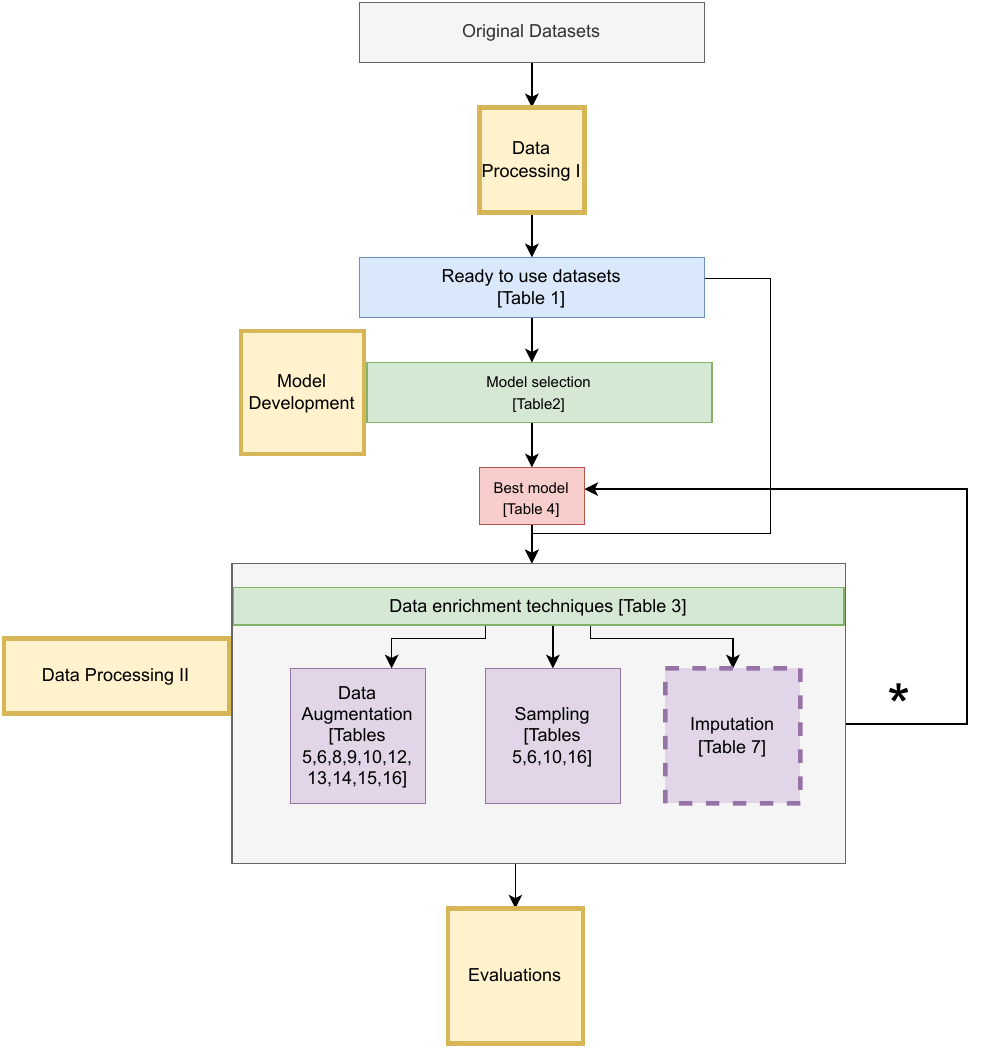}
  \caption{Experimental Setup}
  \label{fig:experimental_setup}
\end{figure}

\begin{figure}[!ht]
  \centering
   \includegraphics[width=0.8\linewidth]{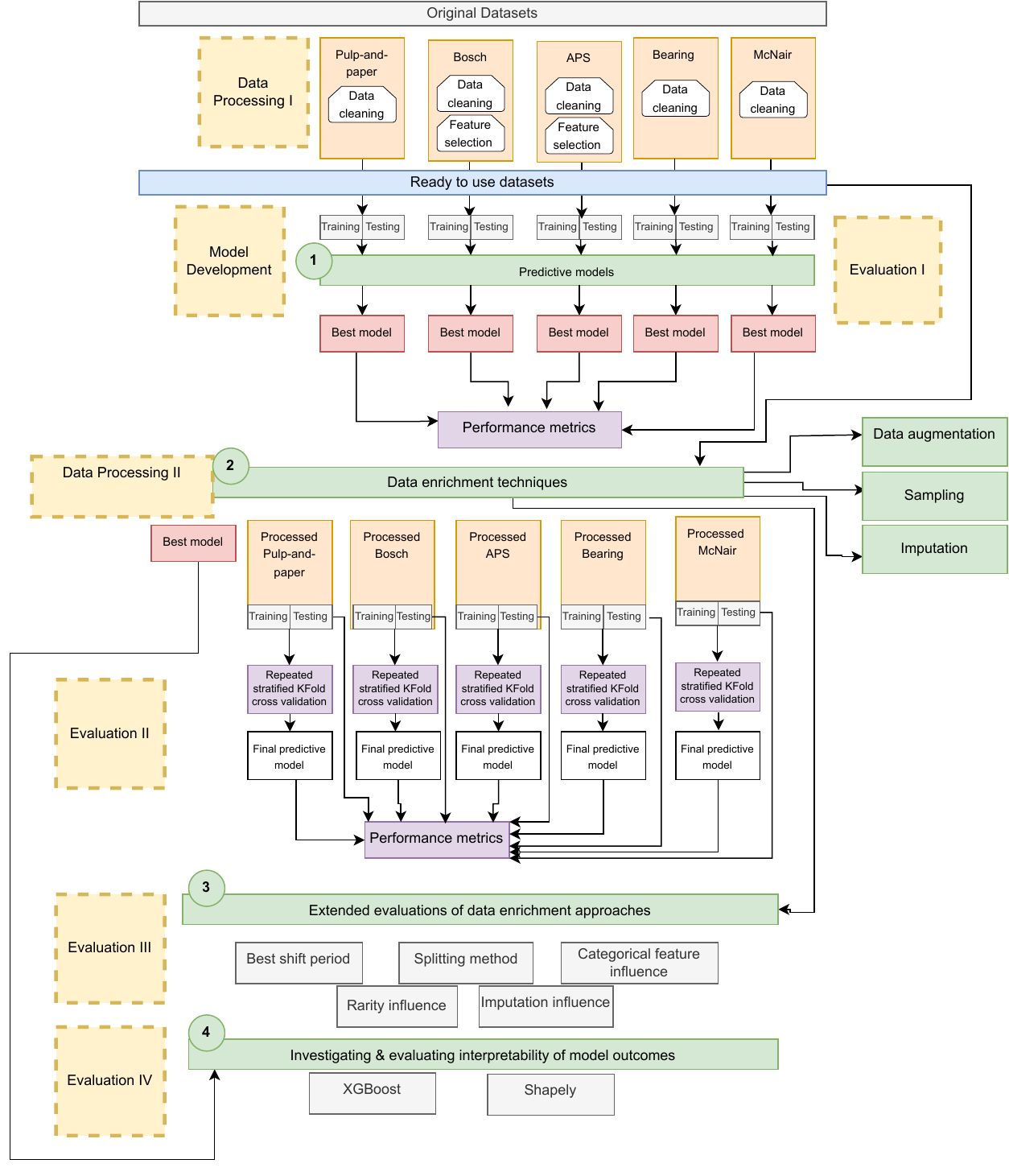}
  \caption{Detailed flow diagram of the steps involved in the proposed experimental setup.}
  \label{fig:overall_summary}
\end{figure}

The main objective of this research is to improve the detection and prediction of rare events in manufacturing by leveraging data enrichment approaches. Figure \ref{fig:experimental_setup} shows a high-level depiction of the proposed experimental framework used in this investigation. The process involves primary data processing, followed by model development and selecting the best model for each dataset. Subsequently, secondary data processing (Data Processing II) is undertaken, incorporating the application of data enrichment techniques, culminating in comprehensive evaluations.   A detailed flow diagram of the steps involved in the experimental setup is shown in Figure \ref{fig:overall_summary}. Next, each of the steps involved in this setup will be elaborated in detail.

\subsection{Datasets}
This research employs multivariate numerical time series data, wherein records encompass the continuous recording of multiple interrelated data streams over a period of time. Considering manufacturing and Industry 4.0 as domains of interest, we select five real-world datasets from different manufacturing sectors. These datasets can be categorized into two distinct types: naturally rare datasets and derived datasets. Naturally rare event datasets refer to datasets that inherently exhibit a low occurrence rate of specific events or phenomena (\cite{shyalika2023comprehensive}). Conversely, a derived dataset results from the transformation of an existing dataset that is initially not rare concerning the events, thereby creating rarity in the derived set (\cite{shyalika2023comprehensive}). As naturally rare datasets, we used the Pulp-and-paper manufacturing dataset (\cite{ranjan2018dataset}), Bosch production line performance dataset (\cite{kaggleBoschProduction}), Air Pressure System (APS) failure dataset (\cite{APSFailureDataSet, uciMachineLearningTrucksAPS}), and the Future Factories (FF) dataset (\cite{harik2024analog, Harik_2024}).  Derived datasets include the Ball-bearing dataset (\cite{case_school_of_engineering_2021}) and a sampled instance of the FF dataset. Among these datasets, only the pulp-and-paper manufacturing dataset and the FF dataset include a timestamp variable within their final datasets. Conversely, the remaining three datasets lack timestamp variables in their final configurations; however, they are organized chronologically. Consequently, the pulp-and-paper manufacturing dataset and the FF dataset were utilized specifically for rare event prediction, whereas all five datasets were experimented with for rare event detection. A detailed overview of these datasets is presented below.

\subsubsection{Pulp-and-paper manufacturing dataset}
The rare event dataset (\cite{ranjan2018dataset}) has been collected from the pulp-and-paper industry and includes data collected from several sensors placed in different parts of the machines in a paper mill along its length and breadth.  These sensors measure metrics on raw materials  (e.g., amount of pulp fiber, chemicals, etc.) and process variables  (e.g., blade type, couch vacuum, rotor speed, etc.). The dataset has 18,398 records collected over 30 days. Data includes sensor readings at regular time-intervals (x’s), which are two minutes apart and the event label (y). The records have the time(timestamp), y: the binary response variable, and x1-x61: predictor variables. All the predictors are continuous variables, except x28 and x61. x61 is a binary variable, and x28 is a categorical variable. There are only 124 rows with y = 1, rest are y = 0. 1 denotes a sheet break. 


\subsubsection{Bosch production line performance dataset}
Bosch created the Bosch production line performance dataset (\cite{kaggleBoschProduction}) in August 2016, which was introduced as part of a Kaggle competition designed to address the detection of defective parts within assembly lines. This dataset represents a complex industrial system characterized by four distinct segments and 52 individual workstations. Each of these workstations conducts a varying number of tests and measurements on specific components, collectively yielding a dataset encompassing a substantial 4264 features. Importantly, the components produced along the assembly line may follow distinct paths, lacking a common initiation or termination workstation. The dataset contains records of 1,183,747 individual parts, each associated with one of 4700 unique combinations. The dataset has been divided into three separate tables for the training and testing sets, each containing specific data types. In these tables, "L" designates the assembly line number, "S" represents the workstation number, and "F" denotes the value of an anonymous feature measured at a given station. Notably, the code "D" is employed uniquely, as columns labeled as "D(n)" record timestamps corresponding to features "F(n - 1)." Notably, a substantial portion of the dataset consists of NULL values, with only 5\% of numeric values, 1\% of categorical values, and 7\% of timestamps being non-null. For the scope of our analysis, we only considered the dataset containing numerical values. The numerical dataset comprises 970 distinct features. To maintain consistency with the number of features in the pulp-and-paper dataset, we employed Random Forest feature importance scoring to identify and select 59 variables that were deemed most significant. 


\subsubsection{Air Pressure System (APS) failure dataset}
The dataset (\cite{APSFailureDataSet}) contains data from Scania trucks, primarily focusing on the Air Pressure System (APS) responsible for tasks like braking and gear changes. It includes two classes: one for APS component failures (positive) and the other for failures not related to APS (negative). The dataset comprises 60,000 examples in the training set (59,000 negative and 1,000 positive) and 16,000 examples in the test set, with 171 attributes per record. The dataset is numeric and the dataset represents a subset of all available data meticulously curated by domain experts. The attribute names are anonymized, consisting of numerical counters and histograms with open-ended conditions. There are 171 attributes in total, with 7 being histogram variables. The dataset has numerous missing values, with 8 attributes having more than 50\% missing values and only 2\% of instances having complete data, while some instances exhibit up to 80\% missing values. Consistent with the approach used in the Bosch dataset, we employed Random Forest feature importance scoring to identify and subsequently select 59 variables, prioritizing those deemed to be most significant. 



\subsubsection{Ball-bearing dataset}

The dataset (\cite{case_school_of_engineering_2021}) from Case Western Reserve University, Cleveland, Ohio, encompasses a range of components and parameters for analyzing ball bearings. The apparatus used to collect these data includes a 2hp motor, a torque transducer/encoder, a dynamometer, and control electronics.  The bearings have three fault diameters of 7 mils, 14 mils, 21 mils, and 28 mils (1 mil=0.001 inches). There are three types of wear: inner raceway wear, outer raceway wear, and ball bearing wear. Data are collected by the two accelerometers placed at the 12 o’clock position at both the drive end and fan end of the apparatus. Data includes normal baseline data, 12 kHz fault data(for drive end), 48kHz fault data(for fan end) collected with a sampling rate of 10s each
and are in Matlab (*.mat) format.

For our research, we employed two primary datasets for analysis. The first dataset, \textit{'Normal\_0.mat'}, comprises samples obtained under normal operating conditions from the normal baseline data. These samples were collected when the motor load was at zero horsepower (HP) and operating at approximately 1797 revolutions per minute (rpm). The faulty dataset utilized for our study includes samples from the outer race 12k Drive End Bearing Fault Data, specifically the dataset labeled \textit{'OR007@6\_0.mat'}. These samples were selected due to an outer race fault and were obtained under similar conditions, with the motor operating at zero HP and an approximate speed of 1797 rpm. The original ball-bearing dataset does not exhibit rarity. Consequently, for our analysis, two distinct rare datasets were generated by maintaining the normal-to-rare ratios at 5\% and 0.5\%, and we used drive-end accelerometer data (DE) and fan-end accelerometer data (FE) as the features. 


\subsubsection{Future Factories (FF) dataset}
The fourth real-world dataset (\cite{harik2024analog, Harik_2024}) used in the study was generated by the McNair Aerospace Center at the University of South Carolina in September 2023, with a specific application in rocket assembly. It originally encompassed data from runs lasting three hours, six hours, 7.5 hours, and 24 hours. Our investigation is focused on the six-hour data run. The comprehensive dataset includes various metrics such as conveyor VFD temperature, conveyor workstation statistics, cycle statistics (cycle count, cycle state, and cycle number), factory cell statistics (cabinet state and door statuses), material handling station statistics (relating to the picking of rocket parts), and data from four robots (load cell measurements, potentiometer data, and six different angles). The analysis within this study centers on two response variables, namely "Success" and "Anomaly," with "Success" chosen as the primary response variable for our investigation. The initial frequency of the FF dataset was 10 Hz. A subsampled dataset was created by down-sampling the original data to a frequency of 1 Hz. 
After careful feature exploration and investigations, 12 features were taken for the modeling phase. 
As the second primary contribution of this paper, we will be introducing the original processed dataset we prepared from the six-hour run and the subsampled dataset.


The statistics of the datasets we used in the study are given in the Table \ref{tab:statistics}

\begin{table}[!ht]
\centering
\scriptsize
\caption{Datasets and their statistics}
\label{tab:statistics}
\begin{tabular}{ccccccc}

\hline
\multicolumn{1}{l}{}        & \textit{P\&P} & \textit{BS} & \textit{APS} & \textit{\begin{tabular}[c]{@{}c@{}}FF-O\end{tabular}} & \textit{BR} & \textit{\begin{tabular}[c]{@{}c@{}}FF-S\end{tabular}} \\ \hline
\textit{Rarity \%}          & 0.67\%        & 0.58\%         & 1.67\%       & 1.70\%                                                              & 0.5\%, 5\%       & 1.70\%                                                             \\
\textit{Original features}  & 61            & 970            & 171          & 12                                                                  & 2                & 12                                                                 \\
\textit{Selected features}  & 59            & 59             & 59           & 12                                                                  & 2                & 12                                                                 \\
\textit{No. of data points} & 18398         & 1183747        & 60000        & 216034                                                              & 20000            & 21604                                                              \\ 

\textit{Data collection period} & 30 days         & not mentioned        & not mentioned        & 6 hours                                                              & not mentioned            & 6 hours                                                              \\ 
\textit{Industry type} & Paper         & Automotive        & Automotive        & Mechanical                                                              & Mechanical            & Mechanical                                                             \\ 
\textit{Data sampled/not} & Not         & Not        & Not        & Not                                                              & Not            & Yes                                                              \\
\textit{Sampling rate} & -         & -        & -        & -                                                              & -            & 1Hz                                                              \\
\textit{Train/test split} & & & 70:30 & &  
\\
\hline
\end{tabular} \\
\footnotesize{$^*$P\&P-pulp-and-paper, BS-Bosch, APS-Air Pressure Systems, BR-Ball bearing, FF-S-Future factories sampled, FF-O-Future factories original}
\end{table}

\subsection{Methods for Rare Event Detection and Prediction} 
Each dataset was modeled using statistical, machine learning, and deep learning techniques for rare event detection and prediction, as detailed in Table \ref{tab:model_categories}. \textit{Detection} involves the identification of events after they occur. \textit{Prediction} is an ahead-of-time prediction of events, in which we used curve-shifting operations to shift the response labels in the data frame to facilitate early prediction of an event. Within the scope of statistical approaches, the Autoregressive Integrated Moving Average (ARIMA) model was employed. Machine Learning methods involved the utilization of Support Vector Machine (SVM), Adaboost, XGBoost, Weighted XGBoost, Random Forest, Logistic Regression, and Weighted Logistic Regression. For Deep Learning, Simple Recurrent Neural Networks (RNN), Convolutional Neural Networks (CNN), Long Short-Term Memory (LSTM), Autoencoder, and LSTM Autoencoder were implemented. Following an evaluation of their outcomes, the best-performing model for each dataset was selected. Subsequently, employing the identical feature set for each of the datasets, the data enrichment process, which is described in subsection \ref{section:data_enrichment}, was conducted. The selected best-performing model was then applied to model these processed datasets.

Presently, there is a growing trend in the time series research community in developing Transformer-based models for time series forecasting. However there are certain limitations in using transformer-based models in time series like loss of time series temporal information, quadratic complexity of sequence length, slow training and inference speed due to the encoder-decoder architecture, and overfitting issues (\cite{zeng2023transformers, Valeriy_Manokhin_2023, lee2023ts}). Considering these limitations, we have not extended our research to explore transformer-based models as a baseline in our study.

\begin{table}[!ht]
\centering
\caption{Model categories and techniques}
\label{tab:model_categories}
\begin{tabular}{ll}
\hline
\textbf{Model Category}                                     & \textbf{Techniques}  \\ \hline
Statistical   & ARIMA                                \\ \hline
\begin{tabular}[c]{@{}l@{}}Machine \\ Learning\end{tabular} & \begin{tabular}[c]{@{}l@{}}SVM, Adaboost, XGBoost, \\ Weighted XGBoost, Random Forest, \\ Logistic Regression, Weighted Logistic Regression\end{tabular} \\ \hline
\begin{tabular}[c]{@{}l@{}}Deep \\ Learning\end{tabular}    & \begin{tabular}[c]{@{}l@{}}Simple RNN, CNN, LSTM, Autoencoder, \\ LSTM Autoencoder\end{tabular}                                                          \\ \hline
\end{tabular}
\end{table}

\subsection{Data Enrichment Approach}
\label{section:data_enrichment}

\begin{figure}[!ht]
  \centering
   \includegraphics[width=0.75\linewidth]{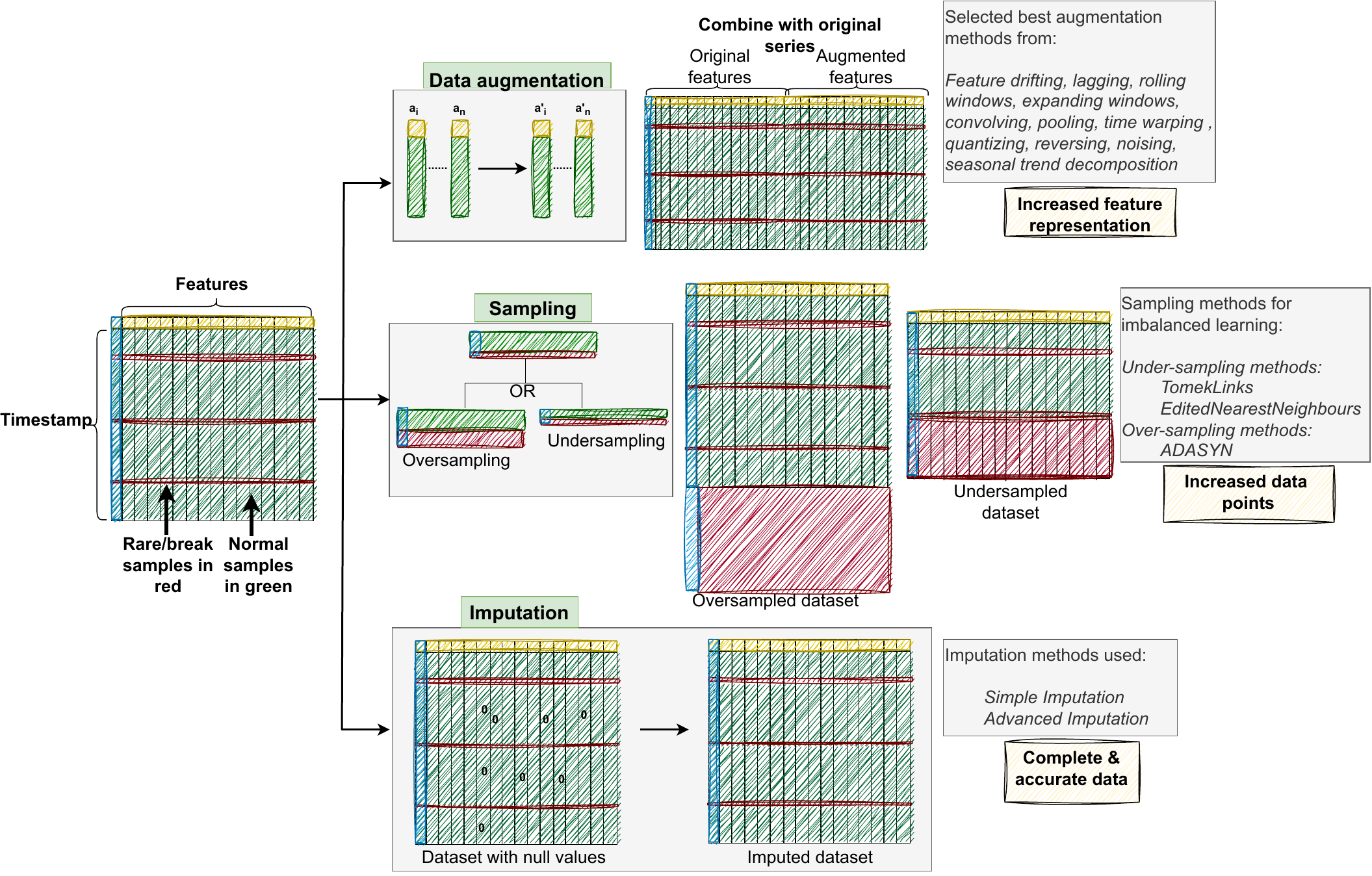}
  \caption{Visual Summary on Data Enrichment Approach }
  \label{fig:data_enrichment_summary}
\end{figure}

Three data enrichment techniques; data augmentation, sampling, and imputation, were executed on the pre-processed datasets as delineated in Table \ref{tab:datatables}. While data augmentation and sampling techniques were applied across all five datasets, imputation was solely performed on the Bosch and APS datasets. The imputation technique was employed to address the substantial volume of missing values inherent in these particular datasets. A visual summary of the data enrichment approaches utilized has been depicted in the Figure \ref{fig:data_enrichment_summary}.

\begin{table}[!ht]
\centering
\caption{Data enrichment approaches performed on different datasets}
\label{tab:datatables}
\begin{tabular}{lccccc}
\hline
\textbf{\begin{tabular}[c]{@{}l@{}}Data enrichment \\ technique\end{tabular}} & \textbf{Pulp} & \textbf{Bosch} & \textbf{APS} & \textbf{Bearing} & \textbf{FF} \\ \hline
Data augmentation                                                             & \checkmark             & \checkmark               & \checkmark             & \checkmark                 & \checkmark                \\
Sampling                                                                      & \checkmark              & \checkmark               & \checkmark             & \checkmark                 & \checkmark                \\
Imputation                                                                    & x             & \checkmark               & \checkmark             & x                & x               \\ \hline
\end{tabular}
\end{table}

\subsubsection{Data Augmentation}
We conducted several basic and advanced time-series-specific data augmentation methods listed below. For each datasets, best-performing and weakly-performing data augmentation methods were identified. In the context of time series forecasting, Basic data augmentation methods encompass fundamental transformations that are employed on existing data to enhance the dataset (\cite{wen2020time}). These methods include techniques like cropping, adding noise, shifting, scaling, or rotating the time series to generate new instances. In contrast, advanced data augmentation methods involve more intricate techniques such as Seasonal Trend Decomposition, Generative Adversarial Networks (GANs), Variational Autoencoders (VAEs), Recurrent GAN (RGAN), etc. (\cite{wen2020time}) applied specifically to generate new synthetic instances while preserving the characteristics and patterns of the original time series data. This section delineates the techniques employed for data augmentation, encompassing both basic and advanced techniques.

\textbf{Basic augmentation techniques}

\begin{enumerate}
    \item Relative change:
Relative change calculation in time series data involves measuring the percentage change between consecutive data points. This calculation provides valuable information about the magnitude and direction of change over time. To calculate relative change, we compared the current value of a data point with its previous value, which is the value of the previous time step. The formula for relative change is as follows:

\begin{equation}
Relative~ change = ((Current~ value - Previous~ value) / Previous~ value) * 100
\end{equation}

The relative change can take positive or negative values, representing an increase or decrease in the data point relative to its previous value. A positive relative change indicates growth or upward movement, while a negative relative change indicates a decline or downward movement. This relative change information provides information on the rate of change and the volatility of the underlying phenomenon captured by the time series. It also helps identify patterns and trends in the time series. Large positive or negative relative changes indicate significant shifts in the data, suggesting the presence of important events or anomalies. By examining these changes over time, we can detect important turning points, abrupt shifts, or gradual trends in the data.

\item Lagged features: 
Lagged features involve incorporating past observations or time lags of the target variable or other relevant features into the dataset. It allows the model to consider the historical behavior or dependencies between variables over time. Lagged features were obtained by shifting the time series data by a certain number of time steps, and the shifted values were used as additional input features. By including lagged features, the model can capture temporal dependencies and exploit the autocorrelation present in the time series data.

\item Rolling window:
The rolling window technique involves creating overlapping windows of a fixed size that move along the time series data. For each window, summary statistics or aggregate features are computed, capturing the local patterns within that window. This technique enables the model to capture short-term dynamics and variations in the data. By sliding the window along the time series, the augmented dataset contains multiple instances with different window positions, providing a broader view of the temporal patterns and facilitating the model's understanding of evolving trends. After empirical experiments for each dataset, we selected the window size, obtained the mean of the values in the window, and used it as a feature across all the input variables.

\item Expanding window:
Expanding window augmentation involves gradually increasing the size of the training data window over time. It starts with a small initial window and progressively expands it to include more historical data. This technique allows the model to learn from the past and progressively incorporate longer-term dependencies. By expanding the window in the series, the augmented dataset was able to encompass a wider time range, enabling the model to capture the overall trends and patterns that develop over a longer duration.

\item Convolving: This involves convolving time series with a kernel window. 1D convolution is used here. Convolution can aid in feature extraction and capture local patterns in time series, allowing the model to discern short-term dependencies and variations. Through convolution, the model can identify significant temporal features, like trends or seasonal patterns.
It also reduces noise by highlighting essential information within the data.

\item Pooling: In pooling, the temporal resolution of the time series can be reduced without changing the length of the series. Pooling can be used as an effective measure to reduce dimensionality by reducing computational time to train time series by reducing parameters and minimizing the chances of overfitting. 

\item Drifting: This includes drifting the value of the time series. The algorithm drifts the value of the time series from its original values randomly and smoothly. The maximal drift and the quantity of drift points determine the extent of drifting. Drifting can help reveal long-term dependencies in time series data.

\item Time warping: Here, the algorithm randomly changes the speed of the timeline. The maximum ratio of maximum or minimum speed and the number of speed changes control the time warping. Hence, time warping introduces variability to time series by modifying the temporal dimensions of the data, and so it enables the generation of diverse time-based sequences. This variation aids in expanding the model's exposure to different temporal patterns within the data, potentially enhancing the robustness and generalization capabilities of the forecasting model.

\item Quantizing: In quantizing, the time series values are controlled to a level set, or the values are rounded to the nearest level in the level set. Quantization can simplify complex data by reducing the number of distinct values. So it can facilitate better pattern recognition and model training.

\item Reversing: This reverses the timeline of the series. Reversing the series allows the model to learn patterns and relationships that might be prevalent in the original series but in reverse order. This technique is particularly significant in situations where there might be a symmetry or cyclic nature within the data that could affect future predictions.

\item Noising: This involves adding random noise to the time series. Random noise is added to every time point of a time series independently and is identically distributed. Noising provides benefits by adding variability to the data, and thereby increasing the model's generalizability, preventing overfitting, and improving the model's resilience to outliers.

\end{enumerate}

\textbf{Advanced augmentation techniques} \\
As an advanced augmentation technique, seasonal trend decomposition analysis was conducted. Seasonal trend decomposition is a technique used to decompose a time series into its underlying components, namely trend, seasonality, and residual. The main points of seasonal trend decomposition are as follows:

\begin{enumerate}
\item
Decomposition: This involves the separation of a time series into its three aforementioned components. This decomposition helps in understanding the underlying patterns and structures in the data, enabling further analysis and modeling.
    \item 
Trend: The trend component represents the long-term patterns of the time series data. It can capture the overall increase or decline in the series over an extended period. 
\item
Seasonality: Seasonality includes the repetitive patterns and fluctuations in a series that occur within a fixed time frame, like daily, weekly, monthly, or yearly cycles. Seasonality can be observed as regular and predictable patterns in the data.
\item
Residual: The residual component represents the random and irregular fluctuations in the data that do not fall under the trend or seasonality components. It includes any unexpected and unpredictable variations, measurement errors, or other factors. The residual can be obtained by subtracting the estimated trend and seasonality from the original data.

\end{enumerate}

\subsubsection{Sampling}

In our study, we experimented with the role of sampling in rare event prediction by employing three sampling techniques.

\begin{enumerate}
    \item 
SMOTE-Tomek links:
The SMOTE-Tomek Links method, initially proposed by Batista et al. in 2003 (\cite{batista2003balancing}), combines the strengths of two techniques: SMOTE (Synthetic Minority Over-sampling Technique) and Tomek Links. SMOTE is capable of generating synthetic data for the minority class, while Tomek Links identifies and removes instances from the majority class that form a Tomek Link with instances from the minority class, thus enhancing the class separation. In SMOTE-Tomek link sampling, firstly, random data points from the minority class are selected. Then, the distance between each chosen data point and its k nearest neighbors is calculated. The difference is then multiplied by a random number between 0 and 1, and the result is added to the minority class as a synthetic sample. This process is repeated until the desired proportion of the minority class is achieved. Secondly, random data points from the majority class are chosen, and if the nearest neighbor of a chosen data point is from the minority class (indicating the presence of a Tomek Link), the Tomek Link is removed.

\item 
SMOTE-ENN:
The SMOTE-ENN method, developed by Batista et al. in 2004 (\cite{batista2004study}), combines the strengths of two techniques: SMOTE (Synthetic Minority Over-sampling Technique) and ENN (Edited Nearest Neighbors). SMOTE is utilized to generate synthetic examples for the minority class, while ENN is employed to eliminate observations from both classes that have a different class label compared to their K-nearest neighbor majority class. In SMOTE-ENN sampling, initially, random data points are selected from the minority class. Then, the distance between each selected data point and its K-nearest neighbors is calculated. The difference is then multiplied by a random number between 0 and 1, and the result is added to the minority class as a synthetic sample. This process continues until the desired proportion of the minority class is attained. Subsequently, the K value, representing the number of nearest neighbors, is determined (default is K=3 if not specified). The K-nearest neighbor of each observation is found among the remaining observations in the dataset, and the majority class label from the K-nearest neighbor is obtained. If the class label of the observation and the majority class label from its K-nearest neighbor is different, the observation and its K-nearest neighbor are removed from the dataset. This process is repeated until the desired proportion of each class is achieved.

\item
ADASYN:
The Adaptive Synthetic (ADASYN) algorithm proposed by He et al. in 2008, (\cite{he2008adasyn}) is an oversampling technique specifically designed for imbalanced datasets that can be used in those with rare events. ADASYN addresses the imbalance by generating synthetic examples for the minority class based on their level of difficulty in learning from the existing data. The algorithm focuses on the minority class instances that are more challenging to classify correctly by assigning them higher weights. It achieves this by estimating the local density distribution of the minority class and generating synthetic samples in regions where the class imbalance is more severe. ADASYN adaptively adjusts the synthetic sample generation process to address the specific characteristics of the dataset, ensuring a better representation of the minority class. By considering local density information, ADASYN can address the class imbalance problem and improve the performance of classifiers.
\end{enumerate}

In the experiments, the above sampling techniques were applied to the augmented dataset and the original dataset to compare the effects of augmentation on the final prediction.

\subsubsection{Imputation}
Two methods were incorporated as imputation techniques.  
Replacing null values with zero was used as a simple imputation technique. An advanced imputation technique that involves rolling mean statistics for time series was proposed.

The steps involved in the advanced imputation method used are as follows.

\begin{enumerate}
    \item 
Compute rolling mean statistics for each column within a specified window size (except for the 'time' column).
\item 
Calculate the means for the current window, the previous window (with a one-time step shift upwards), and the following window (with a one-time step shift downwards).
\item 
Average the means obtained from the previous and subsequent windows and replace null or zero with the average mean. 
\item 
If either null or zero is identified, replace null or zero based on the mean of the previous or subsequent window.
\item 
For the last index in the DataFrame, if a column value is still null, update it with the mean of the previous window. (Note that if the first-row value is null, the initial computation and filling of the rolling mean and substituted values do not cover the first row's missing or null value because the rolling window function requires a certain window size to generate meaningful statistical values. We took this assumption to preserve the temporal dependency of time series)
\end{enumerate}

The section \ref{sec:Methodology} described in detail the methodology followed in the research. The next section will be based on the evaluation of the data enrichment approaches and results. 

\section{Evaluation, Results, and Discussion}

This section presents an evaluation of the developed models in terms of performance measures, utilization within either detection or prediction, an evaluation of the influence of data enrichment techniques, and a comparative analysis of experiments on real manufacturing data to extract dataset-specific novel insights. Finally, it includes an in-depth investigation and evaluation of the interpretability of model outcomes utilizing diverse methods and criteria. The experimental setup we designed can be summarized as follows.

\begin{enumerate}
 
\item Baseline Experiment [Section \ref{sec:baseline}]: Evaluate the baseline performance – Objective: To identify the best-performing method for each dataset in rare event detection and prediction.

\item Primary Experiment [Section \ref{sec:primary}]: Evaluate the effect of data enrichment approaches – Objective: To see how each data augmentation technique improves the overall performance in rare event detection and prediction.

\item Secondary Experiment [Section \ref{sec:secondary}]: Evaluate the dataset-specific features towards model performance  – Objective: To identify the best configuration of input feature representation that yields better predictive performance in rare event detection and prediction.

\item Tertiary Experiment [Section \ref{sec:tertiary}]: Investigating the feature importance and model explainability aspect - Objective: To explain the model predictions through feature importance in rare event prediction.
\end{enumerate}

\subsection{Performance metrics.}

To assess the effectiveness of the best models, various performance metrics were employed, including precision, recall, F1-score, and accuracy. Due to the imbalanced nature of rare events, it is important to note that a higher accuracy score does not necessarily indicate a superior model. These are specifically suitable evaluation metrics in rare event prediction research due to their focus on the minority class, ability to account for imbalanced datasets, trade-off analysis between accuracy and sensitivity, interpretability, and robustness to class distribution changes (\cite{islam2021crash}). Our objective is to develop a model that exhibits a tradeoff between high precision and recall metrics while also achieving a high level of prediction accuracy. 





\subsection{Baseline Experiment: Evaluate the baseline performance}
\label{sec:baseline}

\begin{figure}[!ht]
  \centering
   \includegraphics[width=0.5\linewidth]{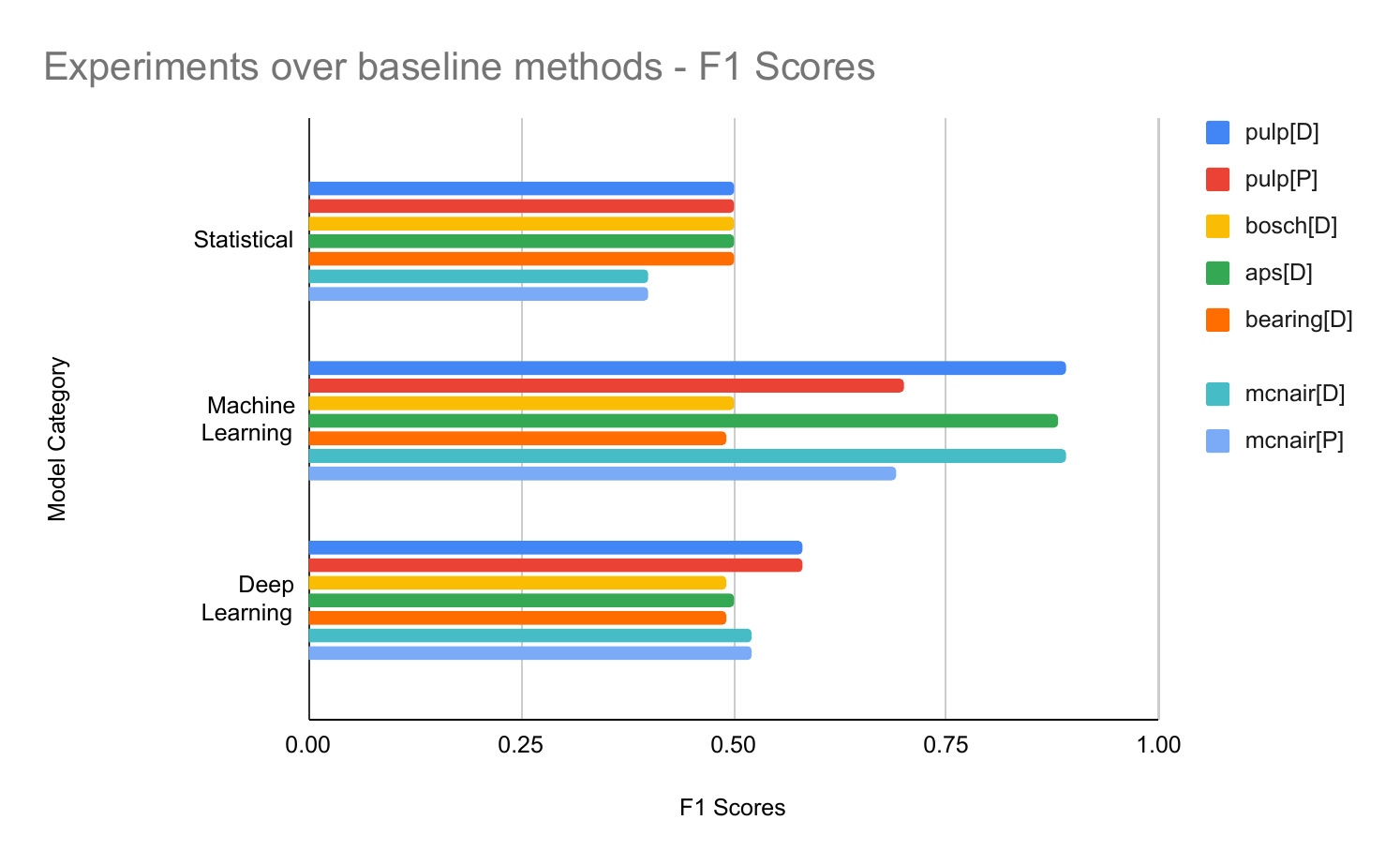}
  \caption{Experiments over baseline methods - F1 Scores}
  \label{fig:F1_alldatasets}
\end{figure}

Table \ref{tab:bestmethods} shows the best modeling technique for each dataset. As per the results, it can be seen that the weighted XGBoost model performed better in all the datasets for both rare event prediction and detection. Figure \ref{fig:F1_alldatasets} shows the highest F1 scores of the baseline models for each model category. 

In all the datasets, several parameters were optimized to maximize the performance of the Weighted XGBoost model and to prevent data overfitting. The optimal XGBoost hyper-parameter values were selected after cross-validation, which are: the number of iterations, max depth, subsample, lambda, alpha, and learning rate. Grid search cross-validation was used in fine-tuning the \textit{scale\_pos\_weight} hyper-parameter, which defines the ratio of the number of samples in negative classes to the positive class. It is a special hyper-parameter designed in XGBoost to tune the behavior of the algorithm for imbalanced classification problems. Repeated stratified KFold cross-validation was used for the parameter tuning, where the performance metrics were optimized.

\begin{table}[!htb]
    \centering
    \caption{Best performing models of all datasets}
    \label{tab:bestmethods}
    \begin{tabular}{lcccc}
        \toprule
        & \textbf{Statistical} & \textbf{ML} & \textbf{DL} \\
        \midrule
        \textbf{Precision} & & & \\
        \quad P\&P[D] & ARIMA (0.5) & XGB (0.89) & CNN (0.60) \\
        \quad P\&P[P] & ARIMA (0.5) & XGB (0.7) & CNN (0.62) \\
        \quad BS[D] & ARIMA (0.5) & XGB (0.51) & Autoencoder (0.53) \\
        \quad APS[D] & ARIMA (0.5) & XGB (0.94) & CNN (0.50) \\
        \quad BR (0.5\%) [D] & ARIMA (0.5) & XGB (0.5) & CNN (0.50) \\
        \quad FF-O[D] & ARIMA (0.4) & XGB (0.89) & Autoencoder (0.52) \\
        \quad FF-O[P] & ARIMA (0.4) & XGB (0.8) & Autoencoder (0.52) \\
        \textbf{Recall} & & & \\
        \quad P\&P[D] & ARIMA (0.5) & XGB (0.89) & CNN (0.56) \\
        \quad P\&P[P] & ARIMA (0.5) & XGB (0.7) & CNN (0.56) \\
        \quad BS[D] & ARIMA (0.5) & XGB (0.50) & Autoencoder (0.54) \\
        \quad APS[D] & ARIMA (0.5) & XGB (0.82) & CNN (0.50) \\
        \quad BR (0.5\%) [D] & ARIMA (0.5) & XGB (0.49) & CNN (0.50) \\
        \quad FF-O[D] & ARIMA (0.4) & XGB (0.89) & Autoencoder (0.54) \\
        \quad FF-O[P] & ARIMA (0.4) & XGB (0.65) & Autoencoder (0.54) \\
        \textbf{F1 Score} & & & \\
        \quad P\&P[D] & ARIMA (0.5) & XGB (0.89) & CNN (0.58) \\
        \quad P\&P[P] & ARIMA (0.5) & XGB (0.7) & CNN (0.58) \\
        \quad BS[D] & ARIMA (0.5) & XGB (0.50) & Autoencoder (0.53) \\
        \quad APS[D] & ARIMA (0.5) & XGB (0.88) & CNN (0.50) \\
        \quad BR (0.5\%) [D] & ARIMA (0.5) & XGB (0.49) & CNN (0.50) \\
        \quad FF-O[P] & ARIMA (0.4) & XGB (0.89) & Autoencoder (0.52) \\
        \quad FF-O[P] & ARIMA (0.4) & XGB (0.69) & Autoencoder (0.52) \\
        \textbf{Support} & & & \\
        \quad P\&P[D,P] & 3655 (3608+47) & & \\
        \quad BS[D] & 236750 (235396+1354) & & \\
        \quad APS[D] & 12000 (11816+184) & & \\
        \quad BR (0.5\%) [D] & 4000 (3983+17) & & \\
        \quad FF-O[D,P] & 42460 (42407+53) & & \\
        \textbf{Accuracy} & & & \\
        \quad P\&P[D] & ARIMA (0.99) & XGB (1) & CNN (0.98) \\
        \quad P\&P[P] & ARIMA (0.99) & XGB (0.98) & CNN (0.98) \\
        \quad BS[D] & ARIMA (0.5) & XGB (0.99) & Autoencoder (0.97) \\
        \quad APS[D] & ARIMA (0.5) & XGB (0.99) & CNN (0.50) \\
        \quad BR (0.5\%) [D] & ARIMA (0.5) & XGB (0.97) & CNN (0.9) \\
        \quad FF-O[D] & ARIMA (0.4) & XGB (0.99) & Autoencoder (0.7) \\
        \quad FF-O[P] & ARIMA (0.4) & XGB (0.95) & Autoencoder (0.79) \\
        \bottomrule
    \end{tabular} \\
\footnotesize{$^*$P\&P-pulp-and-paper, BS-Bosch, APS-Air Pressure Systems, BR-Ball bearing, FF-S-Future factories sampled, FF-O-Future factories original, D-Detection, P-Prediction, ML-Machine learning, DL-Deep learning}
\end{table}

\subsection{Primary Experiment: Evaluate the effect of data enrichment approaches}
\label{sec:primary}

\begin{figure}[!ht]
  \centering
   \includegraphics[width=0.75\linewidth]{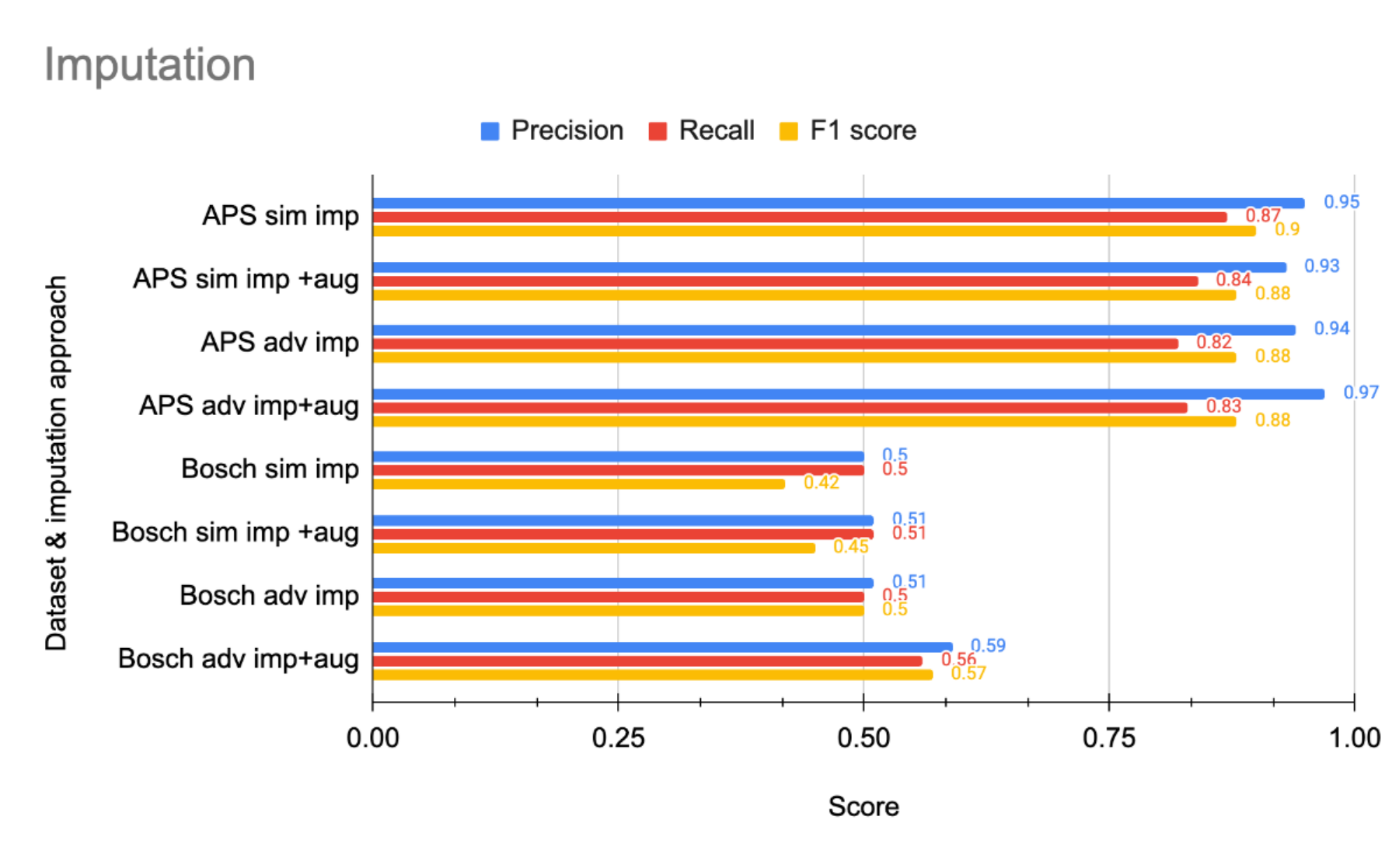}
  \caption{Experiments on simple and advanced imputation with augmentation methods in rare event detection }
  \label{fig:imputation_chart}
\end{figure}

Table \ref{tab:dataenrichment1} in Appendix section \ref{sec:add_exps_detection} and Table \ref{tab:dataenrichment3} show the evaluation results of the best modeling method with and without using data augmentation and sampling techniques in rare event detection and prediction, respectively. In table \ref{tab:dataenrichment4}, we show the best sampling method with and without data augmentation for rare event detection across all datasets. As per the results, it is seen that using augmentation as a data enrichment technique in detection has yielded better weighted average performance in normal and not-normal samples in the pulp-and-paper, APS, bearing, and FF datasets. For the Bosch dataset, using augmentation and Tomek link sampling has given better results. For rare event prediction, data augmentation resulted in high performance in both pulp-and-paper and FF datasets. Table \ref{tab:dataenrichment2} and figure \ref{fig:imputation_chart} show the evaluation results of the best modeling method using simple and advanced imputation methods with data augmentation methods in rare event detection. From the results, it is clear that the advanced imputation method and data augmentation method has resulted in better performance metrics in rare event detection in APS and Bosch datasets.

The outcomes of the experiments reveal that the resampled FF dataset exhibits superior performance, even in the absence of data enrichment techniques. Several factors contribute to this improvement. First, the process of resampling can effectively reduce noise within the dataset, resulting in a cleaner signal for the model to interpret. Second, the reduction of variability in the resampled data contributes to a more stable and consistent dataset, aiding in model generalization. Additionally, the resampling technique enhances the visibility of patterns, enabling the model to focus on more meaningful information and reducing the impact of short-term fluctuations. Finally, the resampling approach proves advantageous in capturing time series seasonality more effectively when operating at a lower frequency. These comprehensive benefits underscore the efficacy of the FF resampling method in enhancing the overall performance of the dataset for predictive modeling purposes.

\begin{table}[]
\centering
\caption{Data augmentation and sampling results across all datasets:Rare event detection}
\label{tab:dataenrichment4}
\begin{tabular}{ll|rrr|rrr|rrr|rrr}
\hline
     &        & \multicolumn{3}{l|}{\begin{tabular}[c]{@{}l@{}}All(aug+\\ tomek \\ samp)\end{tabular}} & \multicolumn{3}{l|}{Aug only}                                           & \multicolumn{3}{l|}{\begin{tabular}[c]{@{}l@{}}Samp only\\ -tomek\end{tabular}} & \multicolumn{3}{l}{No method}                                          \\ \hline
     & Support     & \multicolumn{1}{l}{P}      & \multicolumn{1}{l}{R}      & \multicolumn{1}{l|}{F1}      & \multicolumn{1}{l}{P} & \multicolumn{1}{l}{R} & \multicolumn{1}{l|}{F1} & \multicolumn{1}{l}{P}    & \multicolumn{1}{l}{R}    & \multicolumn{1}{l|}{F1}   & \multicolumn{1}{l}{P} & \multicolumn{1}{l}{R} & \multicolumn{1}{l}{F1} \\ \hline
P\&P & 3655   & .82                        & .94                        & .87                          & \textbf{.92}                   & \textbf{.92}                   & \textbf{.92}                     & .81                      & .92                      & .86                       & .89                   & .89                   & .89                    \\
BS   & 236750 & \textbf{.82}                        & \textbf{.52}                        & \textbf{.53}                        & .59                   & .56                   & \textbf{.57}                     & .5                       & .5                       & .5                        & .51                   & .5                    & .5                     \\
APS  & 12000  & .84                        & .82                        & .83                          & \textbf{.97}                   & .83                   & \textbf{.88}                     & .85                      & .9                       & .87                       & .94                   & .82                   & .88                    \\
BR   & 4000   & .89                        & 1                          & .94                          & \textbf{.97}                   & \textbf{1}                     & \textbf{.97}                     & .5                       & .54                      & .39                       & .5                    & .49                   & .49                    \\
FF-S & 45     & \multicolumn{1}{l}{}       & \multicolumn{1}{l}{}       & \multicolumn{1}{l|}{}        & \multicolumn{1}{l}{}  & \multicolumn{1}{l}{}  & \multicolumn{1}{l|}{}   & \multicolumn{1}{l}{}     & \multicolumn{1}{l}{}     & \multicolumn{1}{l|}{}     & .97                   & .92                   & .94                    \\
FF-O & 42460  & \textbf{1}                          & \textbf{1}                          & \textbf{1}                            & \textbf{.96}                   & \textbf{.99}                   & \textbf{.98}                     & .86                      & .92                      & .89                       & .89                   & .89                   & .89                    \\ \hline
\end{tabular} \\
\footnotesize{$^*$P\&P-pulp-and-paper, BS-Bosch, APS-Air Pressure Systems, BR-Ball bearing, FF-S-Future factories sampled, FF-O-Future factories original,
P-Precision, R-Recall, F1-F1 Score, A-Accuracy, aug-Augmentation, samp- Sampling}
\end{table}

\begin{table}[]
\scriptsize
\centering
\caption{Data augmentation and sampling results across all datasets:Rare event prediction}
\label{tab:dataenrichment3}
\begin{tabular}{llrrrrrrrr}
\toprule
\multicolumn{1}{c}{\begin{tabular}[c]{@{}c@{}}PM\end{tabular}} & \multicolumn{1}{c}{Dataset} & \multicolumn{1}{c}{\begin{tabular}[c]{@{}c@{}}All(aug+\\ tomek \\ samp)\end{tabular}} & \multicolumn{1}{c}{\begin{tabular}[c]{@{}c@{}}All(aug+\\ ENN\\ samp)\end{tabular}} & \multicolumn{1}{c}{\begin{tabular}[c]{@{}c@{}}All(aug+\\ ADASYN \\ samp)\end{tabular}} & \multicolumn{1}{c}{\textbf{\begin{tabular}[c]{@{}c@{}}aug \\ only\end{tabular}}} & \multicolumn{1}{c}{\begin{tabular}[c]{@{}c@{}}samp only-\\ tomek\end{tabular}} & \multicolumn{1}{c}{\begin{tabular}[c]{@{}c@{}}samp only-\\ ENN\end{tabular}} & \multicolumn{1}{c}{\begin{tabular}[c]{@{}c@{}}samp only-\\ ADASYN\end{tabular}} & \multicolumn{1}{c}
{\begin{tabular}[c]{@{}c@{}}No \\ method\end{tabular}} \\
\midrule
 & P\&P & 0.87 & 0.72 & 0.88 & \textbf{0.9} & 0.86 & 0.6 & 0.7 & 0.7 \\
\multirow{-2}{*}{P} & \begin{tabular}[c]{@{}l@{}}FF-O\end{tabular} & \multicolumn{1}{l}{} & \multicolumn{1}{l}{} & \multicolumn{1}{l}{} & \multicolumn{1}{l}{} & \multicolumn{1}{l}{} & \multicolumn{1}{l}{} & \multicolumn{1}{l}{} & \textbf{1} \\
\hline
 & P\&P & 0.72 & 0.74 & 0.7 & \textbf{0.78} & 0.73 & 0.64 & 0.63 & 0.7 \\
\multirow{-2}{*}{R} & \begin{tabular}[c]{@{}l@{}}FF-O\end{tabular} & \textbf{1} & 1 & \textbf{1} & \textbf{0.98} & 0.85 & 0.91 & 0.87 & 0.67 \\
\hline
 & P\&P & 0.78 & 0.73 & 0.76 & \textbf{0.83} & 0.77 & 0.62 & 0.65 & 0.7 \\
\multirow{-2}{*}{F1} & \begin{tabular}[c]{@{}l@{}}FF-O\end{tabular} & \textbf{1} & 0.97 & \textbf{0.99} & \textbf{0.99} & 0.76 & 0.73 & 0.76 & 0.73 \\
\hline
 S & P\&P & \multicolumn{8}{c}{3655 (3608+47)} \\
& \begin{tabular}[c]{@{}l@{}}FF-O\end{tabular} & \multicolumn{8}{c}{42460 (42407+53)} \\
\hline
 & P\&P & 0.99 & 0.99 & 0.99 & \textbf{0.99} & 0.99 & 0.98 & 0.99 & 0.98 \\
\multirow{-2}{*}{A} & \begin{tabular}[c]{@{}l@{}}FF-O\end{tabular} & \textbf{1} & 1 & \textbf{1} & \textbf{1} & 1 & 1 & 1 & 1 \\
\bottomrule
\end{tabular} \\
\footnotesize{$^*$P\&P-pulp-and-paper, BS-Bosch, APS-Air Pressure Systems, BR-Ball bearing, FF-S-Future factories sampled, FF-O-Future factories original,
PM-Performance metric, P-Precision, R-Recall, F1-F1 Score, A-Accuracy, S-Support, aug-Augmentation, samp- Sampling}
\end{table}

\begin{table}[]
\centering
\caption{Imputation influence over data enrichment approach}
\label{tab:dataenrichment2}
\begin{tabular}{llrrrrl}
\toprule
Dataset & \begin{tabular}[c]{@{}l@{}}Imputation \\ method\end{tabular} & \multicolumn{1}{c}{Precision} & \multicolumn{1}{l}{Recall} & \multicolumn{1}{c}{F1 score} & \multicolumn{1}{l}{Accuracy} & Support \\
\midrule
APS & sim. imp. & 0.95 & 0.87 & 0.9 & 0.99 &  \\
 & \begin{tabular}[c]{@{}l@{}}sim. imp.\\ + aug.\end{tabular} & 0.93 & 0.84 & 0.88 & 0.99 &  \\
 & adv. imp., & 0.94 & 0.82 & 0.88 & 0.99 &  \\
 & \textbf{\begin{tabular}[c]{@{}l@{}}adv. imp.\\ + aug.\end{tabular}} & \textbf{0.97} & \textbf{0.83} & \textbf{0.88} & \textbf{0.99} & \multirow{-4}{*}{\begin{tabular}[c]{@{}l@{}}12000\\ (11816+\\ 184)\end{tabular}} \\
 \midrule
Bosch & sim. imp. & 0.5 & 0.5 & 0.42 & 0.7 &  \\
 & \begin{tabular}[c]{@{}l@{}}sim. imp. +\\ aug.\end{tabular} & 0.51 & 0.51 & 0.45 & 0.8 &  \\
 & \textbf{adv. imp.} & \textbf{0.51} & \textbf{0.5} & \textbf{0.5} & \textbf{0.99} &  \\
 & \textbf{\begin{tabular}[c]{@{}l@{}}adv. imp.\\ + aug.\end{tabular}} & \textbf{0.59} & \textbf{0.56} & \textbf{0.57} & \textbf{0.99} & \multirow{-4}{*}{\begin{tabular}[c]{@{}l@{}}236750\\ (235396\\ +1354)\end{tabular}} \\
 \bottomrule
\end{tabular} \\
\footnotesize{$^*$sim. imp.-Simple imputation, adv. imp.-Advanced imputation, aug.- Augmentation}
\end{table}

\subsection{Secondary Experiment: Evaluate the dataset-specific features towards model performance.}
\label{sec:secondary}

We conducted several analysis experiments on the manufacturing datasets aiming to reveal dataset-specific novel insights. These experiments were unique for the dataset and were designed based on factors like dataset-specific features, data generation, rarity influence, and use cases, including detection or prediction, etc.

The unique evaluation criteria designed include the below set of experiments:

\begin{enumerate}
\item 
Best shift period over data enrichment approach: 
We conducted this experiment on the rare event prediction use case based on the pulp-and-paper manufacturing and FF datasets. The aim was to find the best shift period or simply the best ahead-of-time prediction for these datasets. In the pulp-and-paper dataset, since there is a considerably high number of original features(59), experiments were conducted on four subsets of datasets; using original features (59) without any data enrichment approach, using 15 selected features using random forest feature importance, adding data augmentation to those 15 selected features, and using original features (59) with data augmentation techniques. With empirical evaluations with two-minute, four-minute, and six-minute shift periods, it was seen that using original features (59) with data augmentation and four-minute ahead prediction yielded better macro average prediction results of both normal and rare classes in the pulp-and-paper manufacturing dataset, as seen in Table \ref{tab:shift_period_pulp}. In the FF dataset, experiments were conducted on two subsets of datasets; using original features (9) without any data enrichment approach and using original features (9) with data augmentation techniques. With empirical evaluations with one-minute, two-minute, and three-minute shift periods, it was observed that using original features (9) with data augmentation and one-minute ahead prediction yielded better prediction macro average results of both normal and rare classes in the FF dataset as seen in Table \ref{tab:shift_period_next}.

\item 
Splitting method over data  enrichment approach:
This experiment was also conducted on the rare event prediction use case based on the pulp-and-paper manufacturing and FF datasets. The aim was to find the best splitting method that can be used in the best ahead-of-time prediction for these datasets. We analyzed the influence of the splitting method for final prediction over three splitting methods: random splitting, time-based splitting, and run-based splitting. Random splitting involved splitting original time series data into train and test splits randomly based on the ratio 70:30. Time-based splitting involved taking the first 80\% of time as the training set and the rest 20\% as the test set. Run-based splitting involves dividing the split based on the actual running time of the machine. For the pulp-and-paper manufacturing dataset, all three splitting methods were experimented. However, the run-based method was omitted in the FF dataset due to the uninterrupted operation of the data generation machine throughout the data collection process. For the pulp-and-paper manufacturing dataset, four sessions of running the machine were obtained by exploring the dataset, as included in Table \ref{tab:run_stats}. The overall results of this experiment are stated in Table \ref{tab:splitting_methods_pulp} and Table \ref{tab:splitting_methods_next}, and it is seen that using data augmentation methods yielded better performance in all the splitting methods in both the datasets except for one session in the run-based splitting for pulp-and-paper dataset.

\item 
Influence of categorical features over the data enrichment approach:
This experiment evaluated the influence of the categorical feature, \textit{paper types} over data augmentation in rare event prediction in pulp-and-paper dataset on time-based splitting. Time-based splitting was considered to be a good splitting method in this experiment since papers with the same paper types are usually manufactured in the same session. Pulp-and-paper dataset contains six different paper types, which are denoted in the \textit{'x28'} variable in the dataset. Experiments were conducted for the top three most-recorded paper types in the dataset. As per the results in Table \ref{tab:paper_types}, it can be seen that paper types and augmentation do not have a direct association with the performance of predicting a rare event.

\item 
Rarity influence over data enrichment approach:
Ball-bearing dataset was used to create a derived dataset and to experiment with the influence of rarity over data augmentation. Two versions of the dataset, with 0.5\% of rarity and 5\% of rarity, were created for the derived datasets. Given the results in Table \ref{tab:rarity}, we can say that the rarity does not directly affect the results of data augmentation in either case; data augmentation has led to better predictions.

\item 
Best and weak augmentation methods across the datasets:
Best and weak augmentation methods were selected for all five datasets using the \textit{forward selection} feature selection algorithm. Here, we started with having no augmented feature in the model. Then, in each iteration, we kept adding augmented features that best improved the performance of the best model. Those were considered the best augmentation methods. If an addition of an augmented variable did not improve the performance of the model, they were considered weak augmentation methods. The results are shown in Table \ref{tab:forwardselection} in Appendix section \ref{sec:forwardselection}. Here, the pulp-and-paper and FF datasets include results of rare event prediction, whereas bosch, APS, and bearing datasets include results of rare event detection.

\end{enumerate}

\begin{table}[]
\centering
\small
\caption{Shift period vs data augmentation-pulp-and-paper dataset}
\label{tab:shift_period_pulp}
\begin{tabular}{lrrrrrrrrrcllrrr}
\toprule
 & \multicolumn{3}{c}{Precision} & \multicolumn{3}{c}{Recall} & \multicolumn{3}{c}{F1 score} &  \multicolumn{3}{c}{Accuracy} \\
 \midrule
Support & \multicolumn{12}{c}{3655 (3608+47)} \\
 \hline
\begin{tabular}[c]{@{}l@{}}Number of\\ mins ahead\end{tabular} & 2 & 4 & 6 & 2 & 4 & 6 & 2 & 4 & 6  & 2 & 4 & 6 \\
\hline
\begin{tabular}[c]{@{}l@{}}No aug., \\ all features\end{tabular} & 0.5 & 0.7 & 0.83 & 0.5 & 0.7 & 0.6 & 0.5 & 0.7 & 0.65 &  0.99 & 0.98 & 0.98 \\
\hline
\begin{tabular}[c]{@{}l@{}}Selected \\ features\end{tabular} & 0.53 & 0.65 & 0.83 & 0.52 & 0.68 & 0.59 & 0.52 & 0.66 & 0.63 &  0.99 & 0.99 & 0.98 \\
\hline
\begin{tabular}[c]{@{}l@{}}Selected \\features \\ + aug.\end{tabular} & 0.62 & 0.77 & \multicolumn{1}{l}{0.74} & 0.52 & 0.68 & 0.54 & 0.53 & 0.71 & 0.57 & 0.99 & 0.99 & 0.98 \\
\hline \\
\textbf{aug.} & 0.5 & \textbf{0.9} & \textbf{0.9} & 0.5 & \textbf{0.78} & 0.67 & 0.5 & \textbf{0.83} & \textbf{0.73}  & 0.99 & \textbf{0.99} & 0.99 \\
\bottomrule
\end{tabular} \\
\footnotesize{$^*$aug.- Augmentation}
\end{table}

\begin{table}[]
\centering
\caption{Shift period vs data augmentation-FF dataset}
\label{tab:shift_period_next}

\begin{tabular}{lrrrrrrrrrcllrrr}
\toprule
 & \multicolumn{3}{c}{Precision} & \multicolumn{3}{c}{Recall} & \multicolumn{3}{c}{F1 score} & \multicolumn{3}{c}{Accuracy} \\
 \midrule
Support  & \multicolumn{12}{c}{42460(42407+53)} \\
 \hline
\begin{tabular}[c]{@{}l@{}} Number of\\ mins ahead\end{tabular} & 1 & 2 & 3 & 1 & 2 & 3 & 1 & 2 & 3 & 1 & 2 & 3 \\
\hline
\begin{tabular}[c]{@{}l@{}}no aug, \\ all features \\ \end{tabular} & \textbf{0.86} & 0.89 & 0.81 & \textbf{0.67} & 0.64 & 0.63 & \textbf{0.73} & 0.71 & 0.68  & \textbf{1} & 1 & 1 \\
\hline
\textbf{\begin{tabular}[c]{@{}l@{}}all features +\\  aug\end{tabular}} & 1 & \textbf{1} & \textbf{1} & 0.98 & \textbf{0.99} & \textbf{0.99} & 0.99 & \textbf{1} & \textbf{1} &  1 & \textbf{1} & \textbf{1}\\
\bottomrule
\end{tabular}\\
\footnotesize{$^*$aug.- Augmentation}
\end{table}

\begin{table}[]
\centering
\caption{Splitting method vs data augmentation-pulp-and-paper dataset}
\label{tab:splitting_methods_pulp}
\begin{tabular}{llrrrrl}
\toprule
Splitting method & \begin{tabular}[c]{@{}l@{}}w/o aug \\ or w aug\end{tabular} & \multicolumn{1}{c}{Precision} & \multicolumn{1}{l}{Recall} & \multicolumn{1}{c}{F1 score} & \multicolumn{1}{l}{Accuracy} & Support \\
\hline
\multirow{2}{*}{Random} & w/o aug & 0.7 & 0.7 & 0.7 & 0.99 & \begin{tabular}[c]{@{}l@{}}3655\\  (3608\\  +47)\end{tabular} \\
 & \textbf{w aug} & \textbf{0.9} & \textbf{0.78} & \textbf{0.83} & \textbf{0.99} & \begin{tabular}[c]{@{}l@{}}3655\\  (3608\\  +47)\end{tabular} \\
 \hline
\multirow{2}{*}{Time based} & w/o aug & 0.52 & 0.5 & 0.51 & 0.98 & \begin{tabular}[c]{@{}l@{}}3655\\  (3608\\  +47)\end{tabular} \\
 & \textbf{w aug} & \textbf{0.79} & \textbf{0.54} & \textbf{0.57} & \textbf{0.98} & \begin{tabular}[c]{@{}l@{}}3655\\  (3608\\  +47)\end{tabular} \\
 \hline
\multirow{8}{*}{Run based} & \textbf{\begin{tabular}[c]{@{}l@{}}session 1-\\ w/o aug\end{tabular}} & \textbf{1} & \textbf{0.79} & \textbf{0.87} & \textbf{0.99} & \begin{tabular}[c]{@{}l@{}}828 \\ (816+12)\end{tabular} \\
 & \begin{tabular}[c]{@{}l@{}}session 1-\\ w aug\end{tabular} & 0.85 & 0.79 & 0.82 & 0.99 & \begin{tabular}[c]{@{}l@{}}828 \\ (816+12)\end{tabular} \\
 & \begin{tabular}[c]{@{}l@{}}session 2-\\ w/o aug\end{tabular} & 0.81 & 0.66 & 0.71 & 0.99 & \begin{tabular}[c]{@{}l@{}}1458\\ (1442\\ +16)\end{tabular} \\
 & \textbf{\begin{tabular}[c]{@{}l@{}}session 2-\\ w aug\end{tabular}} & \textbf{1} & \textbf{0.75} & \textbf{0.83} & \textbf{0.99} & \begin{tabular}[c]{@{}l@{}}1458\\ (1442\\ +16)\end{tabular} \\
 & \begin{tabular}[c]{@{}l@{}}session 3-\\ w/o aug\end{tabular} & 0.75 & 0.75 & 0.75 & 0.99 & \begin{tabular}[c]{@{}l@{}}823\\ (815+8)\end{tabular} \\
 & \textbf{\begin{tabular}[c]{@{}l@{}}session 3-\\ w aug\end{tabular}} & \textbf{1} & \textbf{0.88} & \textbf{0.93} & \textbf{1} & \begin{tabular}[c]{@{}l@{}}823\\ (815+8)\end{tabular} \\
 & \begin{tabular}[c]{@{}l@{}}session 4-\\ w/o aug\end{tabular} & 0.99 & 0.57 & 0.62 & 0.99 & \begin{tabular}[c]{@{}l@{}}547\\ (540+7)\end{tabular} \\
 & \textbf{\begin{tabular}[c]{@{}l@{}}session 4-\\ w aug\end{tabular}} & \textbf{0.78} & \textbf{0.78} & \textbf{0.78} & \textbf{0.99} & \begin{tabular}[c]{@{}l@{}}547\\ (540+7) \\
 
 \end{tabular} \\
 \bottomrule
\end{tabular}\\
\footnotesize{$^*$w/o aug- without augmentation, w aug- with augmentation}
\end{table}

\begin{table}
\small
\centering
\caption{Run based splitting statistics-pulp-and-paper dataset}
\label{tab:run_stats}
\begin{tblr}{
  hlines,
  hline{1,7} = {-}{0.08em},
  hline{2} = {-}{0.05em},
}
\textbf{Session}       & \textbf{Start time to end time} & {\textbf{Rare event }\\\textbf{count}} & {\textbf{Normal event }\\\textbf{count}} & {\textbf{Delay to start }\\\textbf{next session}} \\
Session 1              & 5/1/99 0:00 to 5/7/99 12:36     & 37                                     & 4140                                     & 52 mins                                           \\
Session 2              & 5/7/99 13:28 to 5/18/99 4:22    & 38                                     & 7288                                     & 13 hrs 54 mins                                    \\
Session 3              & 5/18/99 18:16 to 5/24/99 23:04  & 27                                     & 4114                                     & 1 hr 12 mins                                      \\
Session 4              & 5/25/99 0:16 to 5/29/99 0:06    & 22                                     & 2732                                     & Process ended                                     \\
\textit{Total samples} &                                 & \textit{124}                           & \textit{18274}                           &                                                   
\end{tblr}
\end{table}

\begin{table}[]
\centering
\caption{Splitting method vs data augmentation-FF dataset}
\label{tab:splitting_methods_next}
\begin{tabular}{llrrrrl}
\toprule
Splitting method & \begin{tabular}[c]{@{}l@{}}w/o aug \\ or w aug\end{tabular} & \multicolumn{1}{c}{Precision} & \multicolumn{1}{l}{Recall} & \multicolumn{1}{c}{F1 score} & \multicolumn{1}{l}{Accuracy} & Support \\
\midrule
\multirow{2}{*}{Random} & w/o aug & 0.86 & 0.67 & 0.73 & 1 & \multirow{4}{*}{\begin{tabular}[c]{@{}l@{}}42460 \\ (42407+53)\end{tabular}} \\
 & \textbf{w aug} & \textbf{1} & \textbf{0.98} & \textbf{0.99} & \textbf{1} &  \\
\multirow{2}{*}{Time based} & w/o aug & 0.72 & 0.55 & 0.58 & 1 &  \\
 & \textbf{w aug} & \textbf{0.72} & \textbf{0.7} & \textbf{0.71} & \textbf{1} & \\
 \bottomrule
\end{tabular}\\
\footnotesize{$^*$w/o aug- without augmentation, w aug- with augmentation}
\end{table}

\begin{table}[]
\centering
\caption{Influence of categorical features(ex: paper types) over data enrichment approach - pulp and paper dataset for time-based splitting}
\label{tab:paper_types}
\begin{tabular}{llrrrrl}
\hline
\begin{tabular}[c]{@{}l@{}}Paper \\ type\end{tabular} & \begin{tabular}[c]{@{}l@{}}w/o aug \\ or w aug\end{tabular} & \multicolumn{1}{c}{Precision} & \multicolumn{1}{l}{Recall} & \multicolumn{1}{c}{F1 score} & \multicolumn{1}{l}{Accuracy} & Support                                                                       \\ \hline
\multicolumn{1}{r}{96}                                                        & w/o aug                                                     & 0.49                                                  & 0.5                        & 0.5                                                  & 0.98                         &                                                                               \\
                                                                              & \textbf{w aug}                                              & \textbf{0.99}                                         & \textbf{0.53}              & \textbf{0.56}                                        & \textbf{0.99}                & \multirow{-2}{*}{\begin{tabular}[c]{@{}l@{}}1301\\ (1285+ 16)\end{tabular}} \\ \hline
\multicolumn{1}{r}{82}                                                        & w/o aug                                                     & 0.5                                                   & 0.5                        & 0.5                                                  & 0.99                         &                                                                               \\
                                                                              & w aug                                                       & 0.5                                                   & 0.5                        & 0.5                                                  & 0.99                         & \multirow{-2}{*}{\begin{tabular}[c]{@{}l@{}}876\\ (872+4)\end{tabular}}       \\ \hline
\multicolumn{1}{r}{118}                                                       & w/o aug                                                     & 0.5                                                   & 0.5                        & 0.5                                                  & 0.99                         &                                                                               \\
                                                                              & w aug                                                       & 0.5                                                   & 0.49                       & 0.49                                                 & 0.97                         & \multirow{-2}{*}{\begin{tabular}[c]{@{}l@{}}527\\ (523+4)\end{tabular}}       \\ \hline
\end{tabular} \\
\footnotesize{$^*$w/o aug- without augmentation, w aug- with augmentation}
\end{table}

\begin{table}[]
\centering
\caption{Rarity influence over data enrichment approach-Ball-bearing dataset}
\label{tab:rarity}
\begin{tabular}{rlrrrrl}
\toprule
\multicolumn{1}{l}{Rarity \%} & \begin{tabular}[c]{@{}l@{}}w/o aug \\ or w aug\end{tabular} & \multicolumn{1}{c}{Precision} & \multicolumn{1}{l}{Recall} & \multicolumn{1}{c}{F1 score} & \multicolumn{1}{l}{Accuracy} & Support \\
\midrule
\multirow{2}{*}{5\%} & w/o aug & 0.5 & 0.48 & 0.32 & 0.4 & \multirow{2}{*}{\begin{tabular}[c]{@{}l@{}}4000\\ (3802+17)\end{tabular}} \\
 & \textbf{w aug} & \textbf{0.96} & \textbf{1} & \textbf{0.98} & \textbf{1} &  \\
 \hline
\multirow{2}{*}{0.50\%} & w/o aug & 0.5 & 0.49 & 0.49 & 1 & \multirow{2}{*}{\begin{tabular}[c]{@{}l@{}}4000\\ (3983+17)\end{tabular}} \\
 & \textbf{w aug} & \textbf{0.97} & \textbf{1} & \textbf{0.99} & \textbf{1} & \\
 \bottomrule
\end{tabular}\\
\footnotesize{$^*$w/o aug- without augmentation, w aug- with augmentation}
\end{table}

\subsection{Tertiary Experiment: Investigating the feature importance and model explainability aspect.}
\label{sec:tertiary}

Given the results or outcomes of a detection or prediction, knowing and understanding the conclusion of a model is important. In our study, to understand the modeling behavior, we used two techniques to interpret the results given by the models. We experimented with these two techniques on the rare event prediction task. Firstly, we interpret the model results using XGBoost feature importance technique, and secondly, we use Shapley Additive exPlanations (SHAP) values, proposed by Lundberg and Lee in 2017, (\cite{lundberg2017unified}), to interpret the outputs of the Weighted XGBoost model. This section includes the results of this investigation done on the pulp-and-paper manufacturing dataset.

\begin{enumerate}
    \item 
    \textbf{Model Interpretations using XGBOOST feature importance:}
Here, firstly, a gradient-boosted tree model that includes a plot of the trees was created. Then 
\textit{total cover} feature importance technique in XGBoost algorithm was used to measure the importance or the total coverage of each feature during the construction of the XGBoost model. \textit{Total cover} measure calculates the sum of the number of times a feature is used to split the data across all trees in the ensemble. In XGBoost algorithm, the algorithm builds an ensemble of decision trees, where each tree tries to capture different patterns or relationships in the data. During constructing these trees, the algorithm decides which features to use for splitting the data at each tree node. The 'total cover' importance metric keeps track of the cumulative coverage of each feature across all the trees. A higher 'total cover' value for a feature indicates that the feature has been selected for splitting frequently during the construction of the ensemble. This implies that the feature is considered important in distinguishing or explaining the target variable by the XGBoost model. 
After obtaining the gradient-boosted tree model and total cover measure, we compared the node names of the tree and total cover measure and realized they do match each other. An example for explanation is given in Figure \ref{fig:interpret1}, which includes both the above measures for predicting rare events in pulp-and-paper manufacturing dataset with feature augmentation.

\begin{figure}[!ht]
  \centering
   \includegraphics[width=0.75\linewidth]{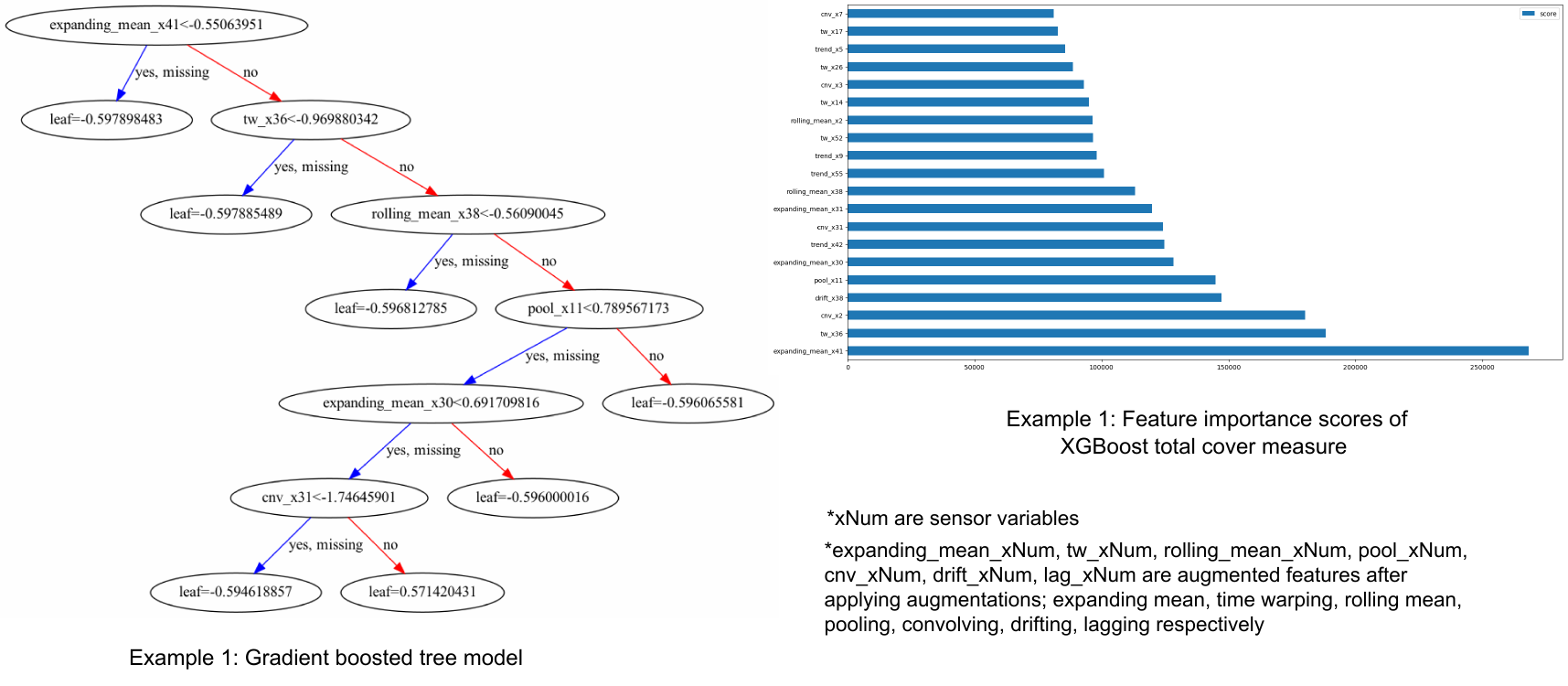}
  \caption{Example:Model Interpretations using XGBOOST feature importance}
  \label{fig:interpret1}
\end{figure}

\item \textbf{Model Interpretations using SHAP - explainable AI with Shapley Additive exPlanations:}
SHAP (SHapley Additive exPlanations) is an explainable AI technique rooted in Shapley values from cooperative game theory. Shapley values represent the mean marginal contribution of individual feature values within all possible values in the feature space. This method provides insight and understanding of how each feature collectively contributes to the outcome predicted by black-box machine learning models. However, it has been researched that SHAP values face limitations such as mathematical inconsistencies, failure to meet human-centric explainability goals, inability to provide causal inference, variable factor contributions, and high computational complexity\cite{kumar2020problems}. We employed three methods using SHAP; Global interpretation, Local interpretation, and Feature dependency analysis, to demonstrate feature analysis. In this section, we present the results of model interpretations based on the data augmentation enrichment technique for the pulp-and-paper manufacturing dataset.

\textbf{i) Global interpretation}
The SHAP values were calculated for every feature (augmented and real) and for the final dataset prepared to understand the feature’s importance and its impact on model output as in the figure \ref{fig:shap1}. Then the features were sorted in descending order based on their SHAP values and their importance in predicting rare events. The X-axis shows the impact of features on the model output. The color represents the average SHAP value of a feature at a position. Red regions have high-value features, and blue regions have low feature values. The vertical width of the color band represents the frequency of a particular SHAP value at a location.
In this dataset, \textit{cnv\_x3} has the greatest impact on the model output. For the majorly visible variables, \textit{cnv\_x3} and \textit{cnv\_x2}, lower values of the feature resulted in higher SHAP values, indicating a lower probability of occurring a break. (negative correlation) When going down in the descending order of feature values, the SHAP values of these features are getting closer to zero, indicating a low impact on the model output. Some variables like \textit{trend\_x9}, \textit{pool\_x11}, \textit{lag\_l\_x3} have a mixed effect on model output. Figure \ref{fig:shap2} presents the average SHAP values for the features across all data samples in the augmented dataset. The average SHAP values for the features are sorted in descending order and show the global importance value of each feature on the model output. The average SHAP values demonstrate that \textit{cnv\_x3}, \textit{cnv\_x2}, \textit{cnv\_x42}, \textit{tw\_x44}, \textit{expanding\_mean\_x47}, \textit{trend\_x29}, \textit{drift\_x23}, \textit{lag\_1\_x19}, \textit{drift\_x42} have the highest impact on the model output.

\begin{figure}[!ht]
  \centering
   \includegraphics[width=0.75\linewidth]{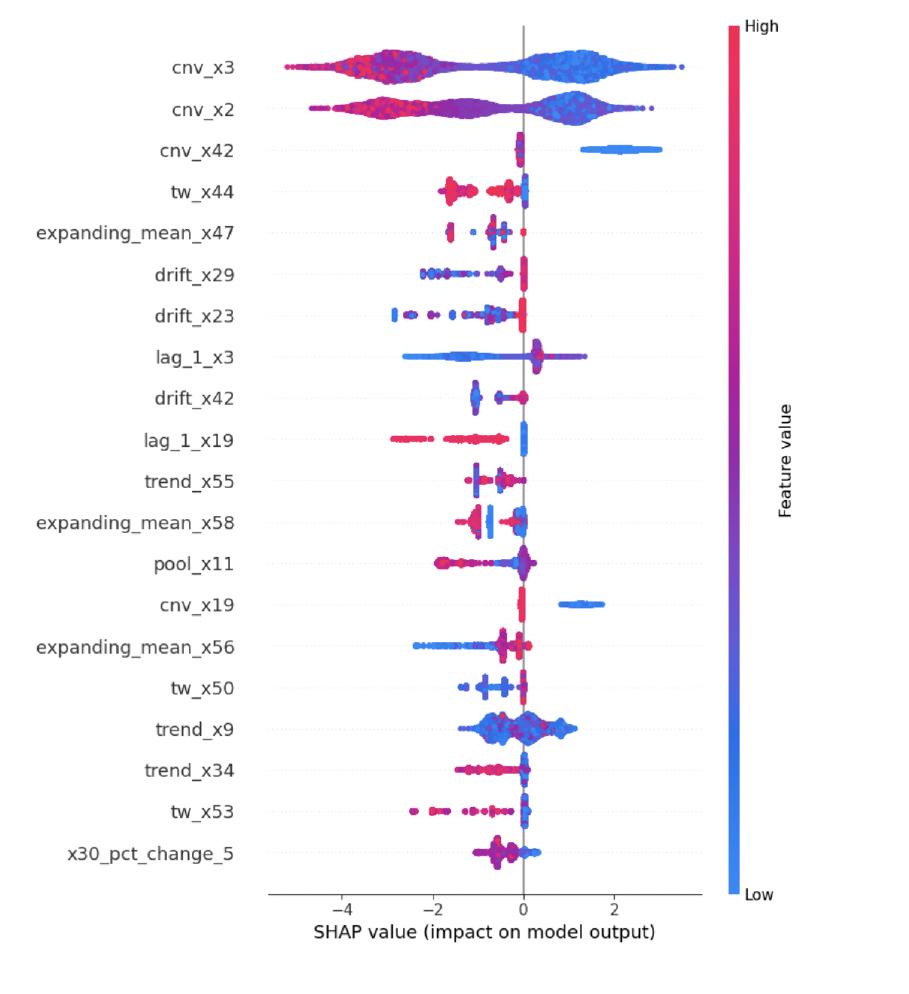}
  \caption{Summary of the SHAP values for pulp-and-paper augmented dataset}
  \label{fig:shap1}
\end{figure}

\begin{figure}[!ht]
  \centering
   \includegraphics[width=0.75\linewidth]{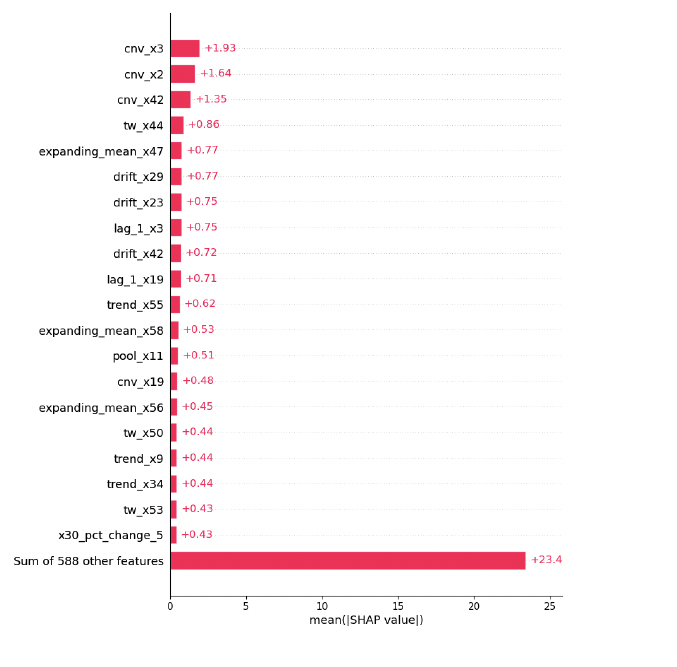}
  \caption{Average SHAP values of the features for pulp-and-paper augmented dataset}
  \label{fig:shap2}
\end{figure}

\textbf{ii) Local interpretation}

\begin{figure}[!ht]
  \centering
   \includegraphics[width=0.75\linewidth]{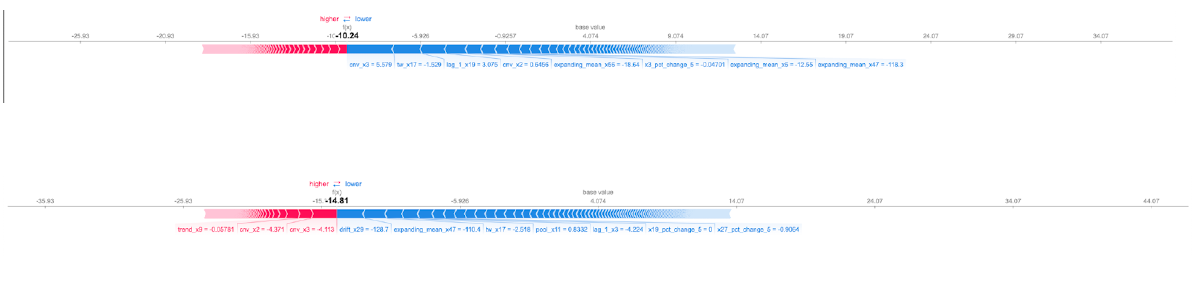}
  \caption{Local effect of the features. (a) Observation 1342, a break event. (b) Observation 14, a normal event.}
  \label{fig:local1}
\end{figure}

The force plot in Figure \ref{fig:local1} is a great visualization to understand the impact of each feature on two selected samples for a given prediction. In this plot, we can see how each feature pushes the model output towards higher or lower values from the base values. In red, we see high Shapley values; in blue, we see low Shapley values. For the positive sample,  we see that the features with the highest Shapley values are \textit{cnv\_x3}, \textit{tw\_x17}, \textit{lag\_1\_x19}, \textit{cnv\_x2}, \textit{expanding\_mean\_x56}. We see a negative correlation between feature values and shapely values. For the negative sample, the features with the lowest Shapley values are \textit{cnv\_x3}, \textit{cnv\_x2}, \textit{trend\_x9}, \textit{drift\_x29}, \textit{expanding\_mean\_x47}, \textit{tw\_x17}, \textit{pool\_x11}. 
Among the features, in \textit{cnv\_x3}, \textit{cnv\_x2}, \textit{trend\_x9}, \textit{drift\_x29}, there is a positive correlation between SHAP values and the feature values, whereas for other features there is a negative correlation between SHAP values and the feature values. In figure \ref{fig:local2}, we show the bar chart and the waterfall plot of the local features corresponding to the two selected samples. In figure \ref{fig:local2}:a(i) and a(ii) we can see that \textit{trend\_x9}, \textit{tw\_x59}, \textit{drift\_x13}, \textit{expanding\_mean\_x3}, \textit{lag\_1\_x3} (those in red) pushed the prediction towards being a break event. On the other hand, we see \textit{cnv\_x3}, \textit{tw\_x17}, \textit{lag\_1\_x19}, and other features in blue colour pushed the prediction towards the normal event.

\begin{figure}[!ht]
  \centering
   \includegraphics[width=0.75\linewidth]{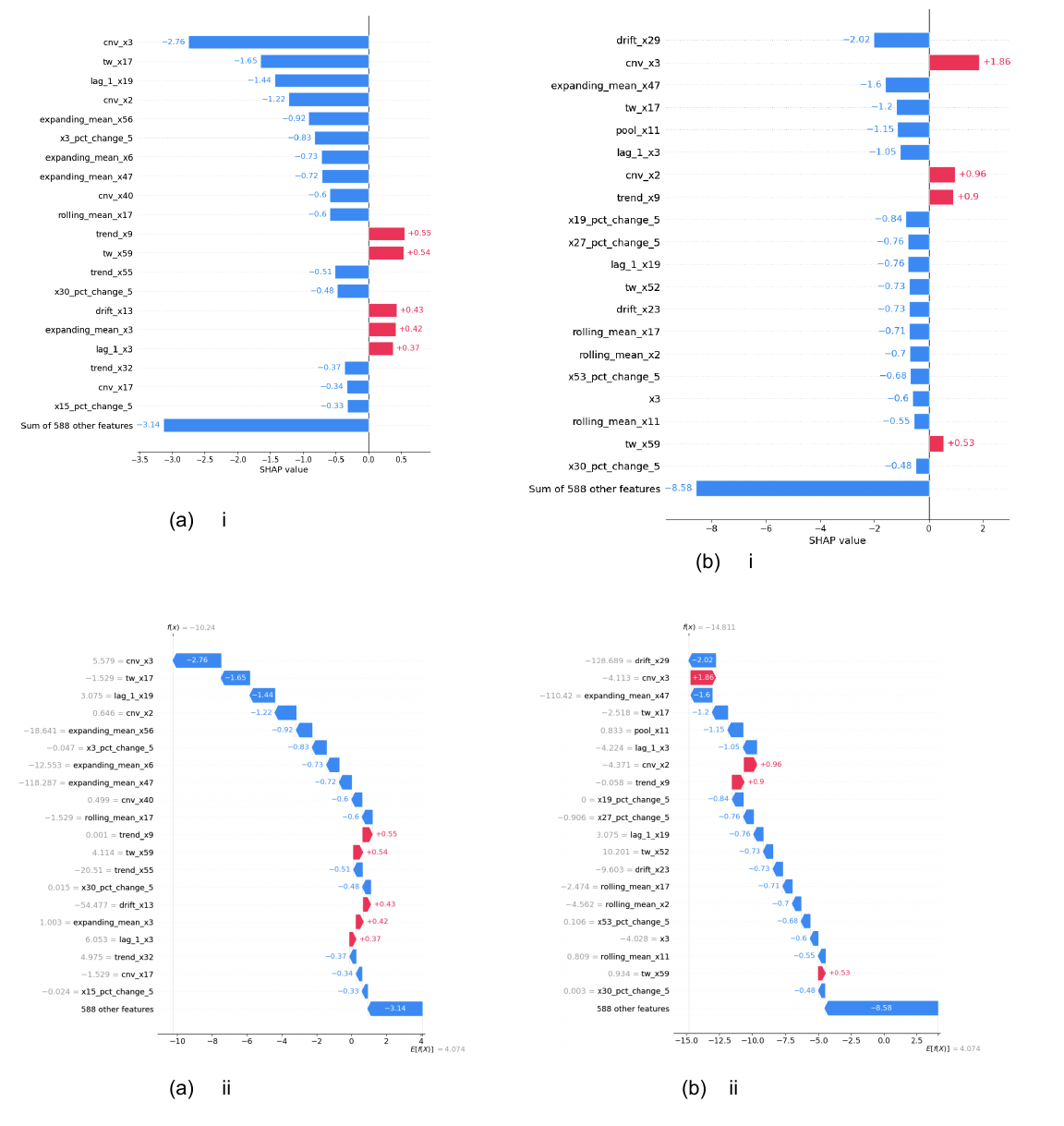}
  \caption{Bar chart(i) and Waterfall plot(ii) of the local effect of the features. (a) Observation 1342, a break event. (b) Observation 14, a normal event.}
  \label{fig:local2}
\end{figure}

For observation b in \ref{fig:local2}:b(i) and b(ii), \textit{cnv\_x3}, \textit{cnv\_x2}, \textit{trend\_x9}, \textit{tw\_x59} (those in red), pushed the prediction towards being a break event. On the other hand, \textit{expanding\_mean\_x47}, \textit{tw\_x17}, \textit{pool\_x11}, \textit{lag\_1\_x3} and other features which are in blue pushed the prediction towards the normal event. Comparison of a and b showed that for both observations, the features \textit{trend\_x9}, \textit{tw\_x59} increase the risk of a break event. 
However, we can see both of those variables having somewhat high percentages for the break class compared with the normal class. For example, the percentage of \textit{tw\_x59} was 4.114\% for the break class; the impact was greater than the normal class, which has a percentage of 0.934\%.

\textbf{iii) Feature dependency analysis}
\\
Feature dependency analysis aims to explore the relationship between the features. A feature dependency plot includes the value of a particular feature on the x-axis and its SHAP value on the y-axis and is generated by changing a specific feature in the model. In the following, we present two examples where feature dependency analysis was analyzed.

\textbf{\textit{(a) Impact of  \textit{cnv\_x3} and  \textit{expanding\_mean\_x17} on model output; }}

In Figure \ref{fig:shap3}:a, \textit{cnv\_x3} was selected as the feature to determine its effect on rare event prediction when the  \textit{expanding\_mean\_x17} variable was changed. The red points represent the higher value of  \textit{expanding\_mean\_x17}, and the blue point represents lower values. It is seen that  \textit{cnv\_x3} contains both positive and negative values, and most negative values have positive SHAP values, and most positive values have negative SHAP values.
So, lower values of the feature resulted in higher SHAP values, thus there’s a negative correlation between  \textit{cnv\_x3} and the probability of a break occurring. For almost all the range of values in  \textit{cnv\_x3} and its SHAP values, we can see  \textit{expanding\_mean\_x17} has high dependence.

\textbf{\textit{(b) Impact of  \textit{tw\_x44} and \textit{trend\_x9} on model output; }}

In Figure \ref{fig:shap3}:b,  \textit{tw\_x44} was selected as the feature to determine its effect on rare event prediction when the  \textit{trend\_x9} variable was changed. The red points represent the higher value of  \textit{trend\_x9}, and the blue point represents lower values. It is seen that  \textit{tw\_x44} contains both positive and negative values in the range 3000 to -4000, and all these values have shap values of zero or negative. It's seen that negative values of  \textit{tw\_x44} having zero shap values and positive values of  \textit{tw\_x44} having negative shap values. Higher values of the feature resulted in lower SHAP values, thus there’s a negative correlation between  \textit{tw\_x44} and the probability of a break occurring. Clearly, there is a dependence between the two variables  \textit{tw\_x44} and \textit{trend\_x9} as seen in b.

\begin{figure}[!ht]
  \centering
   \includegraphics[width=0.75\linewidth]{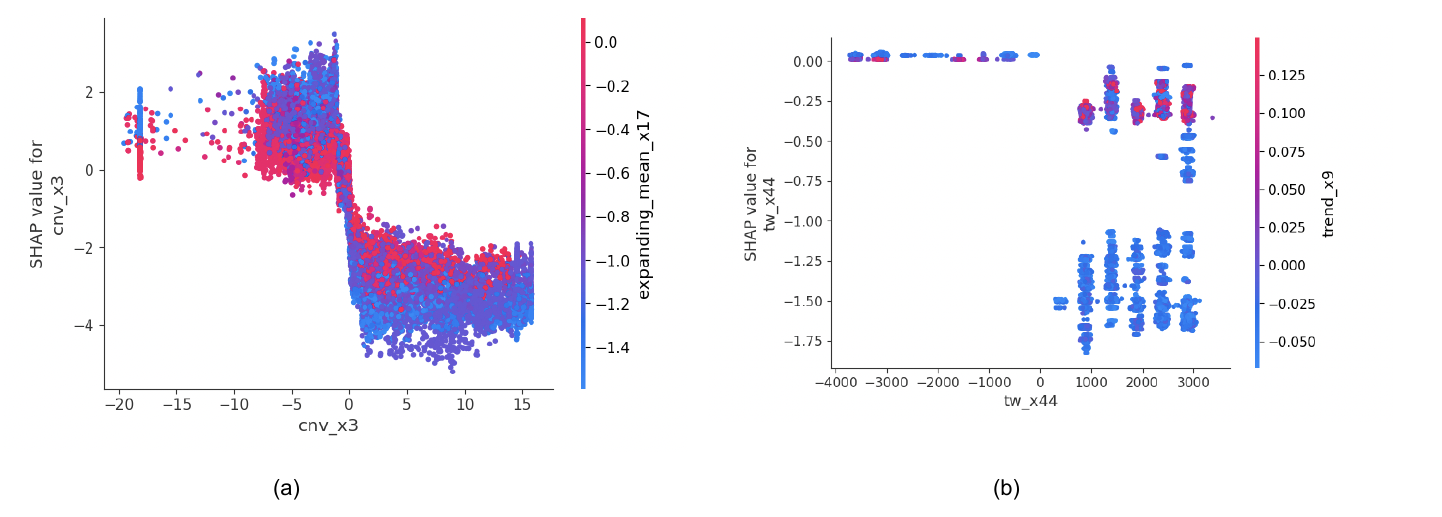}
  \caption{SHAP dependency analysis, (a) Impact of  \textit{cnv\_x3} and  \textit{expanding\_mean\_x17} on model output; (b) Impact of  \textit{tw\_x44} and \textit{trend\_x9} on model output;}
  \label{fig:shap3}
\end{figure}

\end{enumerate}

\newpage
\section{Conclusion}
Detection and prediction of rare events in manufacturing hold the utmost importance in mitigating unintended occurrences and ensuring overall productivity in manufacturing use cases. This study deals with the detection and prediction of rare events in the manufacturing domain by implementing a framework that investigates the influence of data enrichment methods. Based on extensive empirical and ablation experiments on real-world datasets, we gain dataset-specific insights that are important for improving predictive accuracy. Furthermore, examining model interpretability using multiple techniques emphasizes the significance of comprehending the interpretive elements of predictive models in rare-event prediction.  This work will lay the groundwork for future work using more advanced and intelligent approaches to predict rare events that might happen in real-world manufacturing settings.

\newpage
\section{Appendix}
\subsection{Additional experiments on data augmentation and sampling results:Rare event detection [Refer Table {\ref{tab:dataenrichment1}]}}
\label{sec:add_exps_detection}

\begin{table}[!ht]
\centering
\scriptsize
\caption{Data augmentation and sampling results across all datasets:Rare event detection} 
\label{tab:dataenrichment1}
\begin{tabular}{clrrrrrrrr}
\toprule
\begin{tabular}[c]{@{}c@{}}PM\end{tabular} & \multicolumn{1}{c}{Dataset} & \multicolumn{1}{c}{\begin{tabular}[c]{@{}c@{}}All(aug+\\ tomek \\ samp)\end{tabular}} & \multicolumn{1}{c}{\begin{tabular}[c]{@{}c@{}}All(aug+\\ ENN\\ samp)\end{tabular}} & \multicolumn{1}{c}{\begin{tabular}[c]{@{}c@{}}All(aug+\\ ADASYN \\ samp)\end{tabular}} & \multicolumn{1}{c}{\textbf{\begin{tabular}[c]{@{}c@{}}aug \\ only\end{tabular}}} & \multicolumn{1}{c}{\begin{tabular}[c]{@{}c@{}}samp only-\\ tomek\end{tabular}} & \multicolumn{1}{c}{\begin{tabular}[c]{@{}c@{}}samp only-\\ ENN\end{tabular}} & \multicolumn{1}{c}{\begin{tabular}[c]{@{}c@{}}samp only-\\ ADASYN\end{tabular}} & \multicolumn{1}{c}{\begin{tabular}[c]{@{}c@{}}No \\ method\end{tabular}} \\
\midrule
 & P\&P & 0.82 & 0.77 & 0.8 & \textbf{0.92} & 0.81 & 0.81 & 0.8 & 0.89 \\
 & BS & \textbf{0.82} & 0.78 & 0.81 & 0.59 & 0.5 & 0.5 & 0.5 & 0.51 \\
 & APS & 0.84 & 0.76 & 0.74 & \textbf{0.97} & 0.85 & 0.78 & 0.85 & 0.94 \\
 & \begin{tabular}[c]{@{}l@{}}BR-\\(0.5\%)\end{tabular} & 0.89 & 0.87 & 0.87 & \textbf{0.97} & 0.5 & 0.5 & 0.5 & 0.5 \\
 & \begin{tabular}[c]{@{}l@{}}FF-S\end{tabular} & \multicolumn{1}{l}{} & \multicolumn{1}{l}{} & \multicolumn{1}{l}{} & \multicolumn{1}{l}{} & \multicolumn{1}{l}{} & \multicolumn{1}{l}{} & \multicolumn{1}{l}{} & 0.97 \\
\multirow{-6}{*}{P} & \begin{tabular}[c]{@{}l@{}}FF- O\end{tabular} & {\textbf{1}} & 0.99 & \textbf{1} & \textbf{0.96} & 0.86 & 0.79 & 0.85 & 0.89 \\
\midrule
 & P\&P & 0.94 & 0.97 & 0.97 & \textbf{0.92} & 0.92 & 0.97 & 0.92 & 0.89 \\
 & BS & \textbf{0.52} & 0.52 & 0.52 & 0.56 & 0.5 & 0.5 & 0.5 & 0.5 \\
 & APS & 0.82 & 0.89 & \textbf{0.92} & 0.83 & 0.9 & 0.93 & 0.89 & 0.82 \\
 & \begin{tabular}[c]{@{}l@{}}BR-\\(0.5\%)\end{tabular} & 1 & 1 & 1 & \textbf{1} & 0.54 & 0.5 & 0.53 & 0.49 \\
 & \begin{tabular}[c]{@{}l@{}}FF-S\end{tabular} & \multicolumn{1}{l}{} & \multicolumn{1}{l}{} & \multicolumn{1}{l}{} & \multicolumn{1}{l}{} & \multicolumn{1}{l}{} & \multicolumn{1}{l}{} & \multicolumn{1}{l}{} & 0.92 \\
\multirow{-6}{*}{R} & \begin{tabular}[c]{@{}l@{}}FF-O\end{tabular} & {\textbf{1}} & 1 & \textbf{1} & \textbf{0.99} & 0.92 & 0.96 & 0.93 & 0.89 \\
\midrule
 & P\&P & 0.87 & 0.85 & 0.87 & \textbf{0.92} & 0.86 & 0.88 & 0.85 & 0.89 \\
 & BS & \textbf{0.53} & 0.54 & 0.53 & \textbf{0.57} & 0.5 & 0.5 & 0.5 & 0.5 \\
 & APS & 0.83 & 0.81 & 0.8 & \textbf{0.88} & 0.87 & 0.84 & 0.87 & 0.88 \\
 & \begin{tabular}[c]{@{}l@{}}BR-\\(0.5\%)\end{tabular} & 0.94 & 0.92 & 0.92 & \textbf{0.97} & 0.39 & 0.38 & 0.37 & 0.49 \\
 & \begin{tabular}[c]{@{}l@{}}FF-S\end{tabular} & \multicolumn{1}{l}{} & \multicolumn{1}{l}{} & \multicolumn{1}{l}{} & \multicolumn{1}{l}{} & \multicolumn{1}{l}{} & \multicolumn{1}{l}{} & \multicolumn{1}{l}{} & 0.94 \\
\multirow{-6}{*}{F1} & \begin{tabular}[c]{@{}l@{}}FF-O\end{tabular} & {\textbf{1}} & 0.99 & \textbf{1} & \textbf{0.98} & 0.89 & 0.85 & 0.88 & 0.89 \\
\midrule
 & P\&P & \multicolumn{8}{c}{3655 (3608+47)} \\
 & BS & \multicolumn{8}{c}{236750 (235396+ 1354)} \\
 & APS & \multicolumn{8}{c}{12000 (11816 +184)} \\
 & \begin{tabular}[c]{@{}l@{}}BR-\\(0.5\%)\end{tabular} & \multicolumn{8}{c}{4000 (3983 +17)} \\
 & \begin{tabular}[c]{@{}l@{}}FF-S\end{tabular} & \multicolumn{8}{c}{45 (13+32)} \\
\multirow{-6}{*}{\begin{tabular}[c]{@{}c@{}}S \end{tabular}} & \begin{tabular}[c]{@{}l@{}}FF-O\end{tabular} & \multicolumn{8}{c}{42460  (42407+53)} \\
 \midrule
 & P\&P & 1 & 1 & 1 & \textbf{1} & 1 & 1 & 1 & 1 \\
 & BS & 0.99 & 0.99 & 0.99 & \textbf{0.99} & 0.99 & 0.99 & 0.99 & 0.99 \\

 & APS & 0.99 & 0.98 & 0.98 & \textbf{0.99} & 0.99 & 0.99 & 0.99 & 0.99 \\
 & \begin{tabular}[c]{@{}l@{}}BR-\\(0.5\%)\end{tabular} & 1 & 1 & 1 & \textbf{1} & 0.61 & 0.59 & 0.59 & 0.97 \\
 & \begin{tabular}[c]{@{}l@{}}FF-S\end{tabular} & \multicolumn{1}{l}{} & \multicolumn{1}{l}{} & \multicolumn{1}{l}{} & \multicolumn{1}{l}{} & \multicolumn{1}{l}{} & \multicolumn{1}{l}{} & \multicolumn{1}{l}{} & 0.96 \\
\multirow{-6}{*}{A} & \begin{tabular}[c]{@{}l@{}}FF-O\end{tabular} & {\textbf{1}} & 1 & \textbf{1} & \textbf{0.99} & 0.99 & 0.99 & 0.99 & 0.99 \\
\bottomrule
\end{tabular} \\
\footnotesize{$^*$P\&P-pulp-and-paper, BS-Bosch, APS-Air Pressure Systems, BR-Ball bearing, FF-S-Future factories sampled, FF-O-Future factories original,
PM-Performance metric, P-Precision, R-Recall, F1-F1 Score, A-Accuracy, S-Support, D-Detection, P-Prediction, aug-Augmentation, samp- Sampling}
\end{table}

\subsection{Results of data augmentation methods across all datasets [Refer Table {\ref{tab:forwardselection}]}}
\label{sec:forwardselection}
\begin{sidewaystable}[]
\small
\caption{Results of data augmentation methods across all datasets}
\label{tab:forwardselection}
\begin{tabular}{cllllllllllll}
\toprule
\multicolumn{1}{l}{} PM & Dataset & \begin{tabular}[c]{@{}l@{}}Feature \\ drifting\end{tabular} & + lagging & \begin{tabular}[c]{@{}l@{}}+ rolling \\ windows\end{tabular} & \begin{tabular}[c]{@{}l@{}}+ expanding \\ windows\end{tabular} & + quantizing & + convolving & + pooling & + drifting & \begin{tabular}[c]{@{}l@{}}+ time \\ warping\end{tabular} & \begin{tabular}[c]{@{}l@{}}+ STD \\ trend\end{tabular} & \begin{tabular}[c]{@{}l@{}}+STD\\ seasonal\end{tabular} \\
\midrule
\multirow{6}{*}{P} & P\&P {[}P{]} & \textbf{0.75} & \textbf{0.75} & \textbf{0.75} & \textbf{0.75} & \textbf{0.84} & \textbf{0.9} & \textbf{0.93} & \textbf{0.89} & \textbf{0.9} & \textbf{0.9} & 0.9 \\
 & BS {[}D{]} & 0.5 & 0.5 & 0.5 & 0.5 & \textbf{0.84} & \textbf{0.84} & 0.82 & \textbf{0.84} & 0.82 & 0.81 & 0.81 \\
 & APS {[}D{]} & \textbf{0.95} & \textbf{0.51} & \textbf{0.61} & \textbf{0.65} & 0.94 & 0.95 &  &  & \textbf{0.96} & 0.95 & \textbf{0.96} \\
 & \begin{tabular}[c]{@{}l@{}}BR(0.5\%)\\ {[}D{]}\end{tabular} & 0.5 &  &  &  & 0.4 & \textbf{0.7} & \textbf{0.8} & \textbf{0.9} & \textbf{0.95} & 0.5 & \textbf{0.97} \\
 & \begin{tabular}[c]{@{}l@{}}FF-S {[}P{]}\end{tabular} &  & \textbf{0.84} & \textbf{0.89} & \textbf{0.99} &  &  &  &  &  &  &  \\
 & \begin{tabular}[c]{@{}l@{}}FF-O {[}P{]}\end{tabular} & \textbf{0.81} & \textbf{0.69} & \textbf{0.7} & \textbf{0.71} &  &  &  &  &  &  &  \\
\multirow{6}{*}{R} & P\&P {[}P{]} & \textbf{0.68} & 0.5 & 0.5 & 0.5 & \textbf{0.66} & \textbf{0.72} & \textbf{0.72} & \textbf{0.74} & \textbf{0.78} & \textbf{0.78} & 0.77 \\
 & BS {[}D{]} & 0.5 & \textbf{0.5} & \textbf{0.57} & 0.69 & \textbf{0.53} & \textbf{0.53} & 0.52 & \textbf{0.53} & 0.53 & 0.52 & 0.52 \\
 & APS {[}D{]} & \textbf{0.82} &  &  &  & 0.8 & 0.82 &  &  & \textbf{0.83} &  & \textbf{0.84} \\
 & \begin{tabular}[c]{@{}l@{}}BR(0.5\%)\\ {[}D{]}\end{tabular} & 0.49 & \textbf{0.66} & \textbf{0.74} & \textbf{1} & 0.43 & \textbf{0.68} & \textbf{0.8} & \textbf{0.89} & \textbf{0.95} & 0.49 & \textbf{1} \\
 & \begin{tabular}[c]{@{}l@{}}FF-S {[}P{]}\end{tabular} &  & \textbf{0.7} & \textbf{0.72} & \textbf{0.73} &  &  &  &  &  &  &  \\
 & \begin{tabular}[c]{@{}l@{}}FF-O {[}P{]}\end{tabular} & \textbf{0.66} & 0.5 & 0.5 & 0.5 &  &  &  &  &  &  &  \\
\multirow{6}{*}{F1} & P\&P {[}P{]} & \textbf{0.71} & \textbf{0.49} & \textbf{0.6} & \textbf{0.68} & \textbf{0.71} & \textbf{0.79} & \textbf{0.79} & \textbf{0.8} & \textbf{0.83} & \textbf{0.83} & 0.82 \\
 & BS {[}D{]} & 0.5 &  &  &  & \textbf{0.55} & \textbf{0.55} & 0.54 & \textbf{0.55} & 0.55 & 0.54 & 0.54 \\
 & APS {[}D{]} & \textbf{0.87} & \textbf{0.71} & \textbf{0.8} & \textbf{0.99} & 0.8 & 0.87 &  &  & \textbf{0.89} &  & \textbf{0.89} \\
 & \begin{tabular}[c]{@{}l@{}}BR(0.5\%) \\{[}D{]}\end{tabular} & 0.49 & \textbf{0.99} & \textbf{0.99} & \textbf{0.99} & 0.3 & \textbf{0.71} & \textbf{0.8} & \textbf{0.89} & \textbf{0.93} & 0.4 & \textbf{0.97} \\
 & \begin{tabular}[c]{@{}l@{}}FF-S {[}P{]}\end{tabular} &  & 0.99 & 0.99 & 0.99 &  &  &  &  &  &  &  \\
 & \begin{tabular}[c]{@{}l@{}}FF-O {[}P{]}\end{tabular} & \textbf{0.7} & \textbf{0.99} & \textbf{0.99} & 0.99 &  &  &  &  &  &  &  \\
\multirow{6}{*}{S} & P\&P {[}P{]} & \multicolumn{11}{c}{3655 (3608+47)} \\
 & BS {[}D{]} & \multicolumn{11}{c}{236750 (235396+1354)} \\
 & APS {[}D{]} & \multicolumn{11}{c}{12000 (11816+184)} \\
 & \begin{tabular}[c]{@{}l@{}}BR(0.5\%)\\ {[}D{]}\end{tabular} & \multicolumn{11}{c}{4000 (3983+17)} \\
 & \begin{tabular}[c]{@{}l@{}}FF-S {[}P{]}\end{tabular} & \multicolumn{11}{c}{45 (13+32)} \\
 & \begin{tabular}[c]{@{}l@{}}FF-O {[}P{]}\end{tabular} & \multicolumn{11}{c}{42460 (42407+53)} \\
\multirow{6}{*}{A} & P\&P {[}P{]} & \textbf{0.99} & \textbf{0.99} & \textbf{0.99} & \textbf{0.99} & 0.99 & 0.99 & \textbf{0.99} & \textbf{0.99} & \textbf{0.99} & \textbf{0.99} & 0.99 \\
 & BS {[}D{]} & 0.99 & 0.99 & 0.99 & \textbf{0.99} & \textbf{0.99} & \textbf{0.99} & 0.99 & \textbf{0.99} & 0.99 & 0.99 & 0.99 \\
 & APS {[}D{]} & \multicolumn{4}{l}{\textbf{0.99}} & 0.98 & 0.99 &  &  & \textbf{0.99} & 0.99 & \textbf{0.99} \\
 & \begin{tabular}[c]{@{}l@{}}BR(0.5\%)\\ {[}D{]}\end{tabular} & 0.99 & \textbf{0.99} & \textbf{0.99} & 0.99 & 0.99 & \textbf{0.99} & \textbf{0.99} & \textbf{0.99} & \textbf{0.99} & 0.99 & 0.99 \\
 & \begin{tabular}[c]{@{}l@{}}FF-S {[}P{]}\end{tabular} &  & \textbf{0.9}  &\textbf{0.9}  & \textbf{0.9}  &  &  &  &  &  &  &  \\
 & \begin{tabular}[c]{@{}l@{}}FF-O {[}P{]}\end{tabular} & \textbf{0.95} & 0.99 & 0.99 & \textbf{0.99} &  &  &  &  &  &  & \\
 \bottomrule
\end{tabular} \\
\footnotesize{$^*$P\&P-pulp-and-paper, BS-Bosch, APS-Air Pressure Systems, BR-Ball bearing, FF-S-Future factories sampled, FF-O-Future factories original,
PM-Performance metric, P-Precision, R-Recall, F1-F1 Score, A-Accuracy, S-Support, D-Detection, P-Prediction}
\end{sidewaystable}
\newpage

\newpage
\bibliographystyle{unsrtnat}
\bibliography{references}  






\end{document}